\documentclass[]{bytedance_seed}

\usepackage[toc,page,header]{appendix}
\usepackage{booktabs}
\usepackage{multirow}
\usepackage{graphicx}
\usepackage{xcolor}         

\usepackage{minitoc}

\usepackage{booktabs}     
\usepackage{multirow}     
\usepackage{xcolor}       
\usepackage{colortbl}     
\usepackage{graphicx}     

\usepackage{subcaption}
\usepackage{float}

\title{ Balanced Actor Initialization: Stable RLHF Training of Distillation-Based Reasoning Models }

\affiliation[]{ByteDance Seed}
\contribution{Full author list in Contributions}

\abstract{
The development of alignment and reasoning capabilities in large language models has seen remarkable progress through two paradigms: instruction tuning and reinforcement learning from human feedback (RLHF) alignment paradigm, and distillation-based reasoning fine-tuning paradigm. While both approaches prove effective independently, the third paradigm of applying RLHF to distillation-trained models presents significant challenges.  Our investigation reveals two critical phenomena that emerge in this paradigm: Sequence Length Collapse, where language generation dramatically reduces during early RLHF training, and the Reward Hockey Stick Curve, featuring severe reward score drops followed by gradual recovery. These instabilities fundamentally compromise the model's alignment and reasoning capabilities.
To address these challenges, we propose Balanced Actor Initialization (BAI), a two-stage weighted model merging approach. BAI first merges instruction-following and distillation-based reasoning fine-tuned models, then further combines this intermediate model with the pretrained model to preserve foundational knowledge. Through comprehensive experiments across diverse benchmarks and detailed analysis of training experiments, we demonstrate that BAI resolves Sequence Length Collapse, mitigates the Reward Hockey Stick Curve, and enables continuous sequence length improvement during training. Additionally, our analysis reveals that balanced merging ratios achieve optimal trade-offs between training stability and reasoning capability preservation. Our work provides the effective solution for stable training in this third paradigm, enabling more capable reasoning models that combine distillation efficiency with RLHF alignment.
}

\date{\today}
\correspondence{Chen Zheng at \email{chen.zheng1@bytedance.com}}

\begin{document}
\maketitle


\begin{figure}[htbp]
   \centering
   \includegraphics[width=0.7\textwidth,height=160pt]{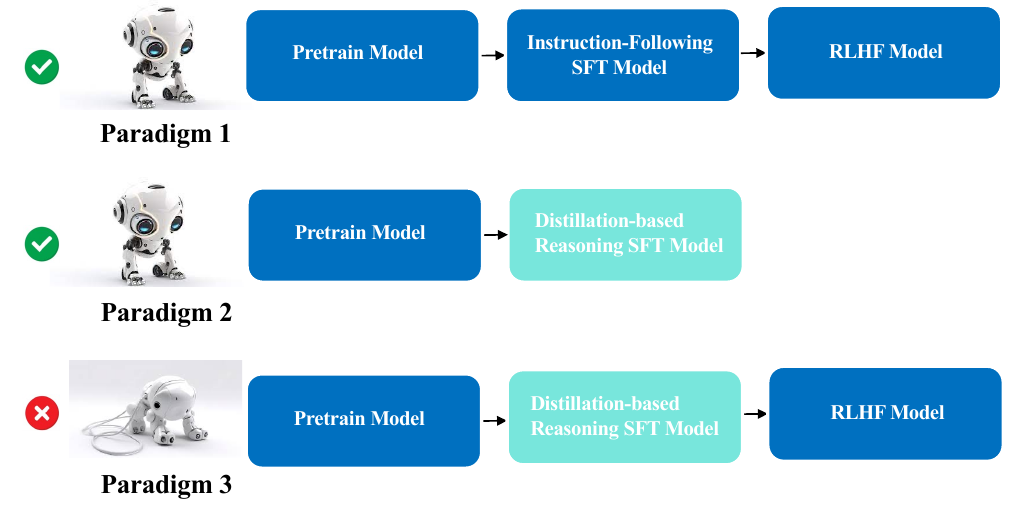}
   \caption{The LLMs training pipeline of Paradigm 1 (instruction tuning and RLHF alignment), Paradigm 2 (distillation-based reasoning fine-tuning), and Paradigm 3 (distillation-based reasoning fine-tuning and RLHF alignment).}
   \label{fig:paradigm}
\end{figure}

\section{Introduction}

The development of alignment and reasoning capabilities in Large Language Models (LLMs) has emerged as one of the most critical challenges in modern artificial intelligence~\cite{openai2024o1,xu2025towards,sui2025stop,liu2025logical}. Recent breakthroughs in chain-of-thought (CoT) reasoning~\cite{wei2022chain,chen2025towards} have demonstrated the potential for models to engage in step-by-step problem solving, leading to significant improvements across diverse reasoning tasks. Moreover, the recent success of DeepSeek-R1~\cite{guo2025deepseek} has demonstrated remarkable capabilities in reasoning and problem-solving, showcasing the potential of advanced post-training methodologies.

Current approaches to developing alignment and reasoning capabilities in language models typically follow two well-established paradigms. As shown in Figure~\ref{fig:paradigm}, \textbf{Paradigm $\textbf{1}$} is the instruction tuning and alignment paradigm, which involves supervised fine-tuning on instruction-following data followed by reinforcement learning from human feedback (RLHF) to align model behavior with human preferences~\cite{ouyang2022training,bai2022training,touvron2023llama,achiam2023gpt,team2023gemini}.
\textbf{Paradigm $\textbf{2}$} is the distillation-based reasoning fine-tuning paradigm, where models are trained on reasoning data distilled from more powerful models~\cite{guo2025deepseek}. This approach enables smaller models that originally lack thinking and reasoning capabilities to acquire sophisticated step-by-step reasoning abilities through supervised learning on distilled data. This paradigm has proven highly effective because distilling reasoning capabilities from larger models into smaller ones yields excellent results with significantly lower computational costs compared to training smaller models through large-scale reinforcement learning~\cite{luo2025deconstructing,li2025llms}. Following the breakthrough of DeepSeek-R1, this paradigm has become increasingly prevalent, with recent works demonstrating substantial performance improvements across various reasoning benchmarks through the use of large quantities of long chain-of-thought data distilled from giant-sized language models.

Given the success of both paradigms independently, a natural question emerges: can we achieve further breakthroughs by combining the instruction tuning and alignment paradigm with distillation-based reasoning fine-tuning? \textbf{Paradigm $\textbf{3}$}—applying RLHF to models that have already undergone distillation-based reasoning fine-tuning-represents a potentially powerful methodology for developing reasoning models that combine the efficiency of distillation with the alignment benefits of human feedback optimization.
However, this third paradigm presents significant challenges. Our experiments reveal that applying RLHF to models trained with extensive distillation-based reasoning fine-tuning leads to critical training instabilities. 

Specifically, models initially generate lengthy reasoning chains after distillation-based reasoning fine-tuning, but after the first several steps of Proximal Policy Optimization (PPO) based reinforcement learning training, the response length experiences a dramatic reduction. We term this phenomenon \textbf{Sequence Length Collapse}. Simultaneously, we observe what we call the \textbf{Reward Hockey Stick Curve}, where reward model scores dramatically drop during early RL training before gradually recovering. These phenomena stem from the fundamental mismatch between specialized reasoning patterns learned during distillation-based fine-tuning and RL optimization requirements, often triggering reward hacking behaviors where models exploit reward signals through shortcuts rather than genuine reasoning improvement. This degradation fundamentally compromises the model's ability to produce detailed reasoning chains and comprehensive responses, representing a critical barrier to successful implementation of the third paradigm.

Recent work~\cite{guo2025deepseek} has shown that incorporating a small amount of cold-start data before reasoning-oriented RL training significantly improves training stability, highlighting the critical importance of robust model initialization. The significance of this cold-start data lies in creating a more balanced model initialization that preserves the model's foundational capabilities while introducing basic reasoning patterns, thereby establishing a stable foundation for subsequent reinforcement learning.
Motivated by this observation, we recognize that robust actor model initialization is essential for addressing the instability issues in the third paradigm. To create such robust initializations and address the Sequence Length Collapse and Reward Hockey Stick Curve phenomena, we propose an effective weighted model merging approach, which we call Balanced Actor Initialization (BAI), that creates robust actor model initializations by combining the pretrained model with instruction-following fine-tuned models and reasoning fine-tuned models at different ratios. This weight merging approach provides a more deterministic and controllable initialization scheme, eliminating the dependence on ambiguous data quantity specifications while offering precise control over the balance between foundational capabilities and reasoning skills.

Specifically, our proposed BAI approach includes two stages to create robust actor initializations. In the first stage, we merge the instruction-following SFT model and the distillation-based reasoning SFT model through weighted linear combination to integrate both instruction-following capabilities and reasoning abilities. 
In the second stage, we further combine this intermediate model from the first stage with the pretrained model to preserve foundational knowledge while maintaining the acquired specialized abilities. 
This two-stage approach directly addresses the challenges of integrating specialized reasoning abilities while preventing the degradation of foundational model capabilities.

Our comprehensive experiments demonstrate that BAI approach successfully resolves Sequence Length Collapse and effectively mitigates the Reward Hockey Stick Curve phenomenon, while providing better control and interpretability. Moreover, by addressing these core instabilities, our simple but effective approach enables the third paradigm to deliver improved performance across diverse evaluation domains. Through extensive BAI ratio experiments, we demonstrate that different merging configurations achieve distinct trade-offs between training stability and alignment and reasoning capability, with balanced ratios demonstrating optimal performance across diverse tasks.

Our paper makes the following key contributions:
\begin{itemize}
   \item We observe and analyze the Sequence Length Collapse phenomenon and Reward Hockey Stick Curve that emerge specifically in the third paradigm when applying RLHF to models trained with extensive distillation-based reasoning fine-tuning, providing comprehensive empirical analysis of their impact on training stability and underlying mechanisms.
   
   \item We propose Balanced Actor Initialization (BAI), a two-stage weighted model merging approach that directly addresses these core instability issues in the third paradigm. BAI resolves Sequence Length Collapse, mitigates the Reward Hockey Stick Curve, and enables continuous sequence length improvement during training, while maintaining model performance across diverse evaluation domains.
   
   \item We demonstrate through extensive experimentation across diverse benchmarks that BAI achieves stable training with improved sequence length maintenance, gradual reward score increases, and enhanced knowledge retention. Through comprehensive ratio analysis, we provide practical guidelines for implementing the third paradigm in alignment and reasoning model development.
\end{itemize}


\section{Related Work}

\subsection{Reinforcement Learning from Human Feedback}


The standard RLHF pipeline consists of three stages: SFT on instruction-following, reward model training using human preference comparisons, and policy optimization using reinforcement learning algorithms. This approach has proven highly effective for improving model helpfulness, harmlessness, and honesty \cite{bai2022constitutional}, leading to the success of models like ChatGPT~\cite{achiam2023gpt}, Claude~\cite{bai2022training}, and Gemini~\cite{team2023gemini}.

Recent advances in RLHF have focused on developing more effective and stable optimization algorithms. Proximal Policy Optimization (PPO) \cite{schulman2017proximal} remains the most widely used approach, providing stable policy updates through clipped objective functions.  Direct Preference Optimization (DPO) \cite{rafailov2024direct} eliminates the need for explicit reward model training by directly optimizing preferences, simplifying the pipeline while maintaining competitive performance. Group Relative Policy Optimization (GRPO) \cite{shao2024deepseekmath} improves sample efficiency by leveraging group-wise preference comparisons. DAPO \cite{yu2025dapo} provides an open-source RLHF system designed for large-scale deployment, while VAPO \cite{yue2025vapo} focuses on efficient and reliable reinforcement learning specifically for advanced reasoning tasks.

The emergence of reasoning-capable models has introduced new challenges and opportunities in RLHF. Recent thinking models demonstrate remarkable capabilities in step-by-step reasoning through explicit chain-of-thought generation. OpenAI's o1-style reasoning models \cite{openai2024o1} pioneered this direction by incorporating sophisticated reasoning protocols into the RLHF framework. Following this breakthrough, DeepSeek-R1 \cite{deepseek2024} demonstrated the successful application of RLHF to reasoning models, showing significant improvements in mathematical and logical reasoning tasks. Subsequently, models like SEED-1.5-Thinking \cite{seed2025seed1} have further advanced the field by developing superb reasoning capabilities through reinforcement learning approaches. 

However, training thinking models presents unique challenges, particularly regarding initialization strategies and training stability. Unlike traditional language models, reasoning models require careful balance between maintaining reasoning capabilities learned during SFT and adapting to human preferences through RL~\cite{zheng2024balancing,zheng2024mistral}. DeepSeek-R1 \cite{deepseek2024} observed that training directly from base models leads to unstable cold start phases, while using a small amount of long CoT data for initialization improves stability. 
Unlike previous approaches that focus on data-based solutions, we propose BAI as a more controllable approach for achieving stable RLHF training while preserving reasoning capabilities.

\subsection{Model Merging}

Model merging has emerged as a powerful technique for combining the capabilities of multiple trained models without requiring additional training data or computational resources. Solar~\cite{kim2023solar} demonstrated that simple weight averaging could improve model performance across different domains. This foundational approach has since been extended to more sophisticated merging strategies.

Recent advances in model merging include task-specific weight interpolation~\cite{matena2022merging,kim2023solar}, where models trained on different tasks are combined to create multi-capable systems. The TIES-Merging approach~\cite{yadav2024ties} addresses sign conflicts and magnitude differences when merging models with overlapping capabilities. Fisher-weighted averaging~\cite{matsuoka2022merging} leverages Fisher information to determine optimal combining weights, while SLERP-based approaches~\cite{goddard2024arcee} use spherical linear interpolation for smoother parameter transitions.

More recently, evolutionary and optimization-based merging methods have gained attention. Recent works~\cite{akiba2024evolutionary,li2025model} introduced algorithms for discovering optimal merging configurations, while DARE~\cite{tang2024dare} addresses redundant parameters during merging. These approaches demonstrate that model merging can achieve performance comparable to or exceeding individual specialized models across diverse tasks.

Different from all these works, which primarily focus on merging models to create multi-capable systems for inference, our work addresses this gap by specifically examining model merging as an effective approach for creating robust initializations for reinforcement learning training, particularly in the context of reasoning model development.
\section{Methodology}

As shown in Figure~\ref{fig:paradigm}, Paradigm $1$ represents the traditional instruction tuning and alignment approach, Paradigm $2$ employs distillation-based reasoning fine-tuning without subsequent RLHF, and Paradigm $3$ combines distillation-based reasoning fine-tuning with RLHF.
However, Paradigm 3 faces critical training instabilities that compromise model performance. These instabilities manifest in two ways: models experience dramatic sequence length reduction during early RL training, losing their ability to generate detailed reasoning chains, while simultaneously exhibiting severe reward model score fluctuations that disrupt the learning process. 


To address the Sequence Length Collapse and Reward Hockey Stick Curve phenomena, we propose Balanced Actor Initialization (BAI), a two-stage weighted model merging approach that creates robust initializations for RL training. BAI combines multiple models to achieve an optimal balance between reasoning capabilities, instruction-following abilities, and foundational knowledge retention.

\subsection{Balanced Actor Initialization (BAI)}
Our approach focuses on creating robust actor model initializations through strategic model merging before RL training begins. The core motivation of BAI is to address the initialization challenges inherent in the third paradigm by leveraging the complementary strengths of different model states while mitigating their individual limitations.
BAI operates in two distinct stages, each addressing different aspects of the initialization challenge. The first stage focuses on capability integration, combining specialized fine-tuned models to create a unified representation of reasoning and instruction-following abilities. Notably, this stage is flexible and can accommodate scenarios where only the distillation-based reasoning model is available, without requiring additional instruction-following components. The second stage emphasizes knowledge preservation, merging the integrated model with the original pretrained model to retain foundational capabilities that are crucial for stable RL optimization. This hierarchical design ensures that the final initialization maintains the delicate balance required for successful training in the third paradigm.

\subsubsection{Multi-SFT Model Merging}

First, given $N$ well-trained SFT models with different capabilities for merging, we denote the parameters of the $i$-th model as $\mathbf{M}_i^{\text{sft}}$ for $i \in \{1, 2, \ldots, N\}$. Each well-trained model is assigned a weighting coefficient $w_i$ that determines its contribution to the final merged model. The merged SFT model $\mathbf{M}_{\text{merge}}^{\text{sft}}$ is then computed as a weighted linear combination:
\begin{equation}
\mathbf{M}_{\text{merge}}^{\text{sft}} = \sum_{i=1}^{N} w_i \mathbf{M}_i^{\text{sft}}
\end{equation}

where the weights sum to one $\sum_{i=1}^{N} w_i = 1$ to preserve parameter scale. This formulation allows for flexible integration of instruction-following and reasoning capabilities from different fine-tuning stages, enabling optimal balance between specialized capabilities while maintaining model stability. In this work, our first stage, Multi-SFT Model Merging, effectively combines the instruction-following SFT model from Paradigm 1 with the distillation-based reasoning SFT model from Paradigm 2 using uniform weights ($w_1 = 0.5$, $w_2 = 0.5$) to retain both strong instruction-following and reasoning capabilities.

\subsubsection{Balanced Model Merging for RL Actor Initialization}

While the merged SFT model $\mathbf{M}_{\text{merge}}^{\text{sft}}$ possesses strong instruction-following capabilities, it often suffers from catastrophic forgetting of the broad knowledge encoded in the original pretrained model. Direct use of such specialized models as RL actors can lead to suboptimal performance due to this knowledge degradation. To address this limitation, we perform a second-stage merging between the merged SFT model and the original pretrained model:
\begin{equation}
\mathbf{M}^{\text{BAI}}  = \alpha \cdot \mathbf{M}^{\text{base}} + \beta \cdot \mathbf{M}_{\text{merge}}^{\text{sft}}
\label{eq:weighted_merge}
\end{equation}

where $\alpha \in [0,1]$ ratio and $\beta=(1-\alpha)$ ratio represent the merging weight that controls the balance between pretrained knowledge and instruction-following capabilities, $\mathbf{M}_{\text{merge}}^{\text{sft}}$ is the merged SFT model from the previous step, $\mathbf{M}^{\text{base}}$ is the original pretrained model, and $\mathbf{M}^{\text{BAI}}$ serves as our proposed RL actor initialization.


The two-stage design addresses key challenges in the third paradigm: (1) Stage 1 integrates complementary capabilities from different fine-tuning approaches while maintaining parameter compatibility; (2) Stage 2 preserves the rich factual knowledge and linguistic capabilities of the pretrained model, preventing catastrophic forgetting that commonly occurs during intensive fine-tuning; (3) The parameterized control through $\alpha$ and $\beta$ provides interpretable trade-off management and control between knowledge retention and behavioral adaptation.

\section{Experiments and Analysis}

\subsection{Implementation Details}

In this work, we conducted RLHF experiments on Seed-MoE-2.5B/25B models. For PPO-based RLHF, we used AdamW as the optimizer, setting both the actor model and critic model learning rates to $1 \times 10^{-6}$. The learning rate employed a warmup-constant scheduler. The global batch size was $4096$, with each prompt sampled once, and the mini-batch size set to $512$. The actor model was initialized using our BAI approach. The critic model was initialized using a reward model, with the GAE $\lambda$ set to $0.95$ and $\gamma$ set to $1.0$.
The training utilized distributed computing across $8$ nodes with a total of $64$ GPUs. The training incorporated advanced optimization techniques including Megatron~\cite{shoeybi2019megatron} parallelism and flash attention~\cite{dao2022flashattention}, etc. 
Most RL experiments were trained for $1600$ steps, except for Paradigm 3 without BAI and Paradigm 3 with BAI ($\alpha=0.6, \beta=0.4$ merging ratio), which were trained for $3000$ steps.

\subsection{Benchmarks}

To comprehensively evaluate BAI's effectiveness, we conduct experiments using Seed-MoE-2.5B/25B across multiple benchmarks spanning knowledge comprehension, reasoning, and conversational tasks.

\textbf{MMLU}~\cite{hendrycks2020measuring}: A widely-adopted benchmark consisting of 15,908 multiple-choice questions spanning 57 subjects from elementary mathematics to advanced professional domains.

\textbf{MMLU-Pro}~\cite{wang2024mmlu}: An enhanced version of MMLU that expands the choice set from four to ten options and integrates more challenging reasoning-focused questions.

\textbf{SuperGPQA}~\cite{du2025supergpqa}: A comprehensive question-answering benchmark that evaluates models' ability to handle complex, multi-step reasoning across diverse knowledge domains with challenging real-world scenarios.

\textbf{LiveBench}~\cite{white2024livebench}: A dynamic, contamination-resistant benchmark with frequently-updated questions from recent information sources and monthly updates.

\textbf{MixEval-Hard}~\cite{ni2024mixeval}: A benchmark that bridges real-world user queries with ground-truth evaluation, achieving 0.96 correlation with Chatbot Arena rankings.

\textbf{ArenaHard (Gemini as Judge)}~\cite{li2024crowdsourced}: A benchmark derived from live Chatbot Arena interactions, consisting of 500 challenging prompts that provide 3× higher model separation compared to MT-Bench.

\textbf{AIME 2024}~\cite{aime2024}: The American Invitational Mathematics Examination, featuring 15 high-difficulty competition mathematics problems that require advanced problem-solving skills and mathematical reasoning.

\textbf{MATH}~\cite{hendrycks2021measuring}: A comprehensive mathematical reasoning benchmark containing 12,500 challenging competition mathematics problems from various domains including algebra, geometry, and number theory.

\textbf{MBPP+}~\cite{austin2021program}: An enhanced version of the Mostly Basic Python Problems benchmark, containing programming challenges that evaluate code generation and algorithmic reasoning capabilities.

\subsection{Performance Comparison Across Paradigms}
We evaluate the effectiveness of our BAI approach by comparing it against the three paradigms capabilities in language models. 
All RLHF experiment evaluations are conducted based on the $1600$-step checkpoint to ensure fair comparison.
Table~\ref{tab:performance_overall} presents comprehensive evaluation results across diverse benchmarks spanning knowledge reasoning (MMLU Pro, MMLU), question answering (SuperGPQA, LiveBench, MixEval-Hard), conversational ability (ArenaHard), mathematical reasoning (AIME $2024$, MATH), and code generation (MBPP+). BAI demonstrates superior performance compared to all three paradigms, achieving the highest overall score of $55.2$, representing a significant improvement over the best individual paradigm (Paradigm $2$ at $53.6$).

The ArenaHard results reveal particularly interesting patterns when using Gemini as the judge. Paradigms $1$ and $3$ receive notably low scores ($15.4$ and $16.0$ respectively), while Paradigm $2$ and BAI achieve substantially higher scores ($34.6$ and $35.9$). This disparity suggests that judgments tend to favor models that exhibit clear reasoning patterns and coherent response generation. Paradigm $1$, lacking extensive reasoning training, produces responses that appear less structured to the judge. Paradigm $3$, despite having reasoning capabilities, suffers from the sequence length collapse and training instabilities that compromise response quality and coherence. In contrast, Paradigm $2$ maintains stable reasoning patterns, while BAI not only preserves these patterns but enhances them through balanced initialization, resulting in the highest ArenaHard score.

BAI shows notable improvements in mathematical reasoning tasks, achieving the highest scores on both AIME 2024 and MATH benchmarks compared to all other paradigms. These improvements demonstrate that BAI successfully preserves and enhances reasoning capabilities while avoiding the degradation typically observed in Paradigm 3. Across most benchmarks, BAI either matches or exceeds the performance of individual paradigms, achieving the highest scores in these benchmarks, indicating the robustness and generalizability of our approach across different task domains.

\begin{table}[]
\centering
\footnotesize
\setlength{\tabcolsep}{4pt}
\renewcommand{\arraystretch}{1.2} 
\begin{tabular}{@{}l@{\hspace{8pt}}*{9}{c}@{\hspace{8pt}}c@{}}
\toprule
\multirow{2}{*}{\textbf{Method}} & \multicolumn{9}{c}{\textbf{Benchmarks}} & \multirow{2}{*}{\textbf{Overall}} \\
\cmidrule(lr){2-10}
 & \rotatebox{45}{\textbf{MMLU Pro}} & \rotatebox{45}{\textbf{MMLU}} & \rotatebox{45}{\textbf{SuperGPQA}} & \rotatebox{45}{\textbf{LiveBench}} & \rotatebox{45}{\textbf{MixEval-Hard}} & \rotatebox{45}{\textbf{ArenaHard}} & \rotatebox{45}{\textbf{AIME 2024}} & \rotatebox{45}{\textbf{MATH}} & \rotatebox{45}{\textbf{MBPP+}} & \\
\midrule
Paradigm 1 & 67.8 & 80.8 & 38.1 & 44.4 & 48.5 & 15.4 & 18.7 & 80.9 & 67.5 & 51.3 \\
Paradigm 2 & 69.7 & 82.0 & \cellcolor{pink!30}\textbf{40.7} & 42.3 & 50.0 & 34.6 & 17.3 & 77.6 & 67.5 & 53.6 \\
Paradigm 3 & 69.2 & 80.8 & 40.5 & 43.0 & 51.5 & 16.0 & 17.7 & 77.5 & 67.2 & 51.5 \\
\midrule
\textbf{BAI} & \cellcolor{blue!25}\textbf{70.2} & \cellcolor{blue!25}\textbf{82.7} & 40.6 & \cellcolor{blue!25}\textbf{44.9} & \cellcolor{blue!25}\textbf{50.8} & \cellcolor{blue!25}\textbf{35.9} & \cellcolor{blue!25}\textbf{21.3} & \cellcolor{blue!25}\textbf{81.0} & \cellcolor{blue!25}\textbf{69.3} & \cellcolor{blue!25}\textbf{55.2} \\
\bottomrule
\end{tabular}
\caption{Performance comparison across different paradigms and the proposed BAI approach. Best results are highlighted in \textcolor{blue}{\textbf{blue}} (BAI) and \textcolor{pink}{\textbf{pink}} (other methods).}
\label{tab:performance_overall}
\end{table}

\subsection{Comprehensive Performance Evaluation Across BAI Configurations}

Table~\ref{tab:performance_ratio} presents the evaluation of performance across different BAI merging configurations, revealing key insights into the optimal balance between pretrained knowledge and specialized reasoning capabilities. The analysis demonstrates a clear trend: configurations with higher SFT weights (lower $\alpha$ values) consistently achieve superior performance across most benchmarks.
The superior performance of SFT-heavy configurations stems from their preservation of reasoning capabilities and instruction-following behaviors acquired during distillation-based reasoning fine-tuning. This advantage is particularly pronounced in benchmarks such as MMLU Pro (70.2\%), MMLU (82.7\%), and ArenaHard (35.9\%), which demand structured problem-solving approaches that align closely with chain-of-thought methodologies. The consistent top-tier performance of the $(\alpha = 0.1, \beta = 0.9)$ configuration across diverse evaluation metrics underscores the effectiveness of prioritizing specialized reasoning patterns over raw pretrained knowledge.

Moreover, configurations with higher pretrain weights ($\alpha \geq 0.6$) also demonstrate competitive performance on specific tasks. For instance, certain benchmarks like MixEval-Hard achieve peak performance at $(\alpha = 0.7, \beta = 0.3)$ (51.2\%), suggesting that the broad knowledge base from pretraining remains valuable for tasks requiring extensive factual recall and general linguistic competence. This task-dependent behavior indicates that the optimal merging ratio may vary based on the specific cognitive demands of different evaluation scenarios.

The ArenaHard benchmark reveals the most dramatic sensitivity to merging ratios, with performance declining precipitously from 35.9\% $(\alpha = 0.1, \beta = 0.9)$ to 11.5\% $(\alpha = 0.9, \beta = 0.1)$. This steep degradation highlights the fundamental importance of instruction-following and conversational reasoning capabilities that are primarily encoded in the SFT component. The results suggest that while pretrained knowledge provides a foundation, the specialized behavioral patterns learned during distillation-based reasoning fine-tuning are indispensable for complex interactive reasoning tasks in the third paradigm.

\begin{table}[]
\centering
\resizebox{\textwidth}{!}{%
\begin{tabular}{@{}lccccccccc@{}}
\toprule
\multicolumn{1}{c}{\multirow{2}{*}{$ \alpha \cdot \mathbf{M}^{\text{base}} + \beta \cdot \mathbf{M}_{\text{merge}}^{\text{sft}} $}} & $\alpha=0.1$        & $\alpha=0.2$ & $\alpha=0.3$ & $\alpha=0.4$ & $\alpha=0.5$ & $\alpha=0.6$ & $\alpha=0.7$ & $\alpha=0.8$ & $\alpha=0.9$ \\
\multicolumn{1}{c}{}                                                                                    & $\beta=0.9$         & $\beta=0.8$  & $\beta=0.7$  & $\beta=0.6$  & $\beta=0.5$  & $\beta=0.4$  & $\beta=0.3$  & $\beta=0.2$  & $\beta=0.1$  \\ \midrule
MMLU Pro                                                                                                & 70.2                & 69.7         & 69.8         & 69.4         & \cellcolor{blue!30}\textbf{70.7}         & 68.3         & 68.6         & 68.2         & 68.1         \\
MMLU                                                                                                    & \cellcolor{blue!25}\textbf{82.7}               & 81.3         & 81.1         & 82.1         & 80.9         & 81.7         & 81.8         & 80.2         & 81.3         \\
SuperGPQA                                                                                               & 40.6                & \cellcolor{blue!30}\textbf{40.8}         & 39.0         & 40.2         & 39.6         & 39.4         & 37.6         & 38.7         & 38.9         \\
LiveBench                                                                                               & 44.9               & 44.6         & 44.0         & 44.3         & 45.0         & 44.6         & \cellcolor{blue!30}\textbf{45.5}         & 44.8         & 44.3         \\
MixEval-Hard                                                                                            & 50.8                & 50.6         & 50.8         & 49.9         & 48.7         & 51.0         & \cellcolor{blue!30}\textbf{51.2}         & 50.0         & 47.8         \\
ArenaHard                             & \cellcolor{blue!25}\textbf{35.9} & 33.3          & 29.9         & 27.0         & 25.2         & 20.7         & 16.0         & 20.3          & 11.5          \\

\bottomrule
\end{tabular}
}
\caption{Performance evaluation across different merging ratios. Best results are highlighted in \textcolor{blue}{\textbf{blue}} color.}
\label{tab:performance_ratio}
\end{table}

\subsection{Reward Hockey Stick Curve Phenomenon in Paradigm 3}

Paradigm 3 faces a critical challenge in the form of reward model instability during RLHF training. When models undergo extensive distillation-based reasoning fine-tuning and are subsequently used as both the reward model and RL actor, we observe the Reward Hockey Stick Curve phenomenon characterized by a "Hockey Stick"-shaped trajectory in reward scores.

\begin{figure}[htbp]
   \centering
   \includegraphics[width=0.9\textwidth]{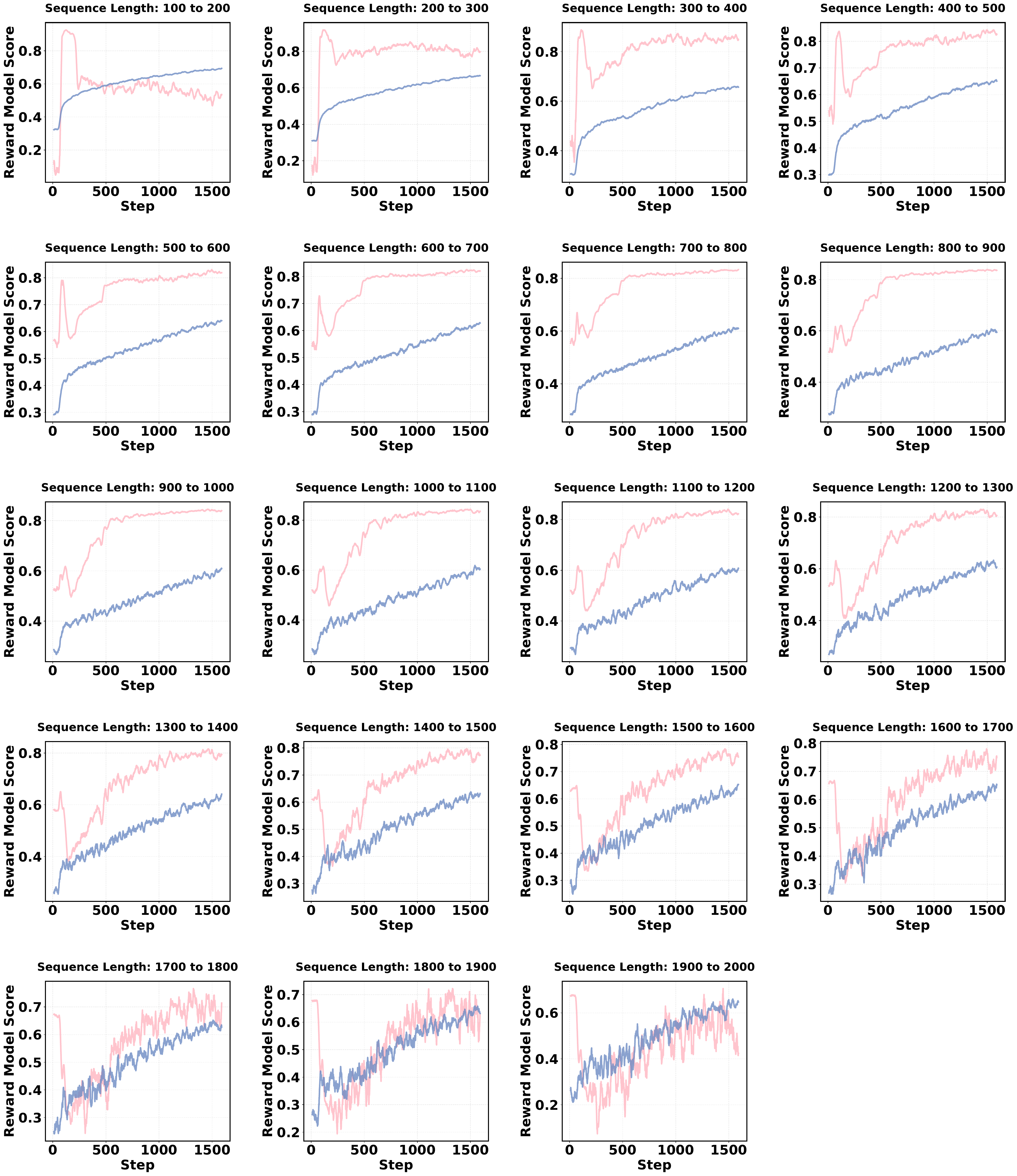}
   \caption{Reward Hockey Stick Curve phenomenon for different generated sequence length. The pink curve represents Paradigm 3 without BAI, while the blue curve represents Paradigm 3 with BAI. Each figure represents the reward scores of samples with generated lengths in the interval from i×100 to (i+1)×100, where i represents integers ranging from 0 to 19.}
   \label{fig:nike_curve_mitigation}
\end{figure}

Figure~\ref{fig:nike_curve_mitigation} illustrates this phenomenon across different sequence length ranges during RLHF training. As shown by the pink curves, the Reward Hockey Stick Curve exhibits three distinct phases: an initial decline where RM scores decrease from their starting values after distillation-based reasoning fine-tuning, a trough phase where performance plateaus at the lowest point, and a recovery phase featuring gradual improvement that often surpasses the original performance levels.

This phenomenon stems from the fundamental mismatch between the specialized reasoning patterns learned during distillation-based fine-tuning and the requirements of RL optimization. The intensive parameter updates during reasoning fine-tuning create optimization landscapes that are highly sensitive to distribution shifts, leading to reward signal instability when transitioning from supervised to RL-generated samples. This mismatch often triggers reward hacking behaviors, where the model learns to exploit the reward signal in unintended ways, initially achieving higher scores through shortcuts rather than genuine reasoning improvement. This instability undermines training consistency and represents a core barrier to successful implementation of the third paradigm.

The Hockey Stick Curve phenomenon directly motivates our Balanced Actor Initialization approach. By creating more robust initial states through weighted model merging, BAI addresses the underlying causes of reward instability and reward hacking tendencies, enabling stable training dynamics from the onset of RL training.

\subsubsection{Mitigating the Reward Hockey Stick Curve through BAI}

To address the Reward Hockey Stick Curve phenomenon in the third paradigm, we demonstrate how our BAI approach effectively mitigates this critical training instability. To validate our method's effectiveness, we employ a balanced $(\alpha=0.5,\beta=0.5)$ merging ratio between the pretrained model and the distillation-based reasoning SFT model, deliberately chosen to demonstrate robustness without extensive hyperparameter optimization.

As shown by the blue curves in Figure~\ref{fig:nike_curve_mitigation}, our BAI approach substantially mitigates the Reward Hockey Stick Curve phenomenon across all sequence lengths, maintaining stable reward trajectories from the onset of training.
The effectiveness of our approach stems from several key factors: BAI creates a more balanced parameter distribution that reduces extreme specialization, preserves broad knowledge from pretraining that enhances stability, exhibits more stable gradient flows during early RL training, and reduces overfitting to specific reasoning patterns. Importantly, the balanced initialization effectively mitigates reward hacking behaviors by providing more robust starting points that are less susceptible to exploiting reward signal shortcuts. The consistent performance across different sequence length ranges indicates that BAI addresses fundamental training dynamics rather than superficial symptoms, while the success of this simple ratio demonstrates the method's practical viability.

\begin{figure}[htbp]
   \centering
   \includegraphics[width=0.9\textwidth]{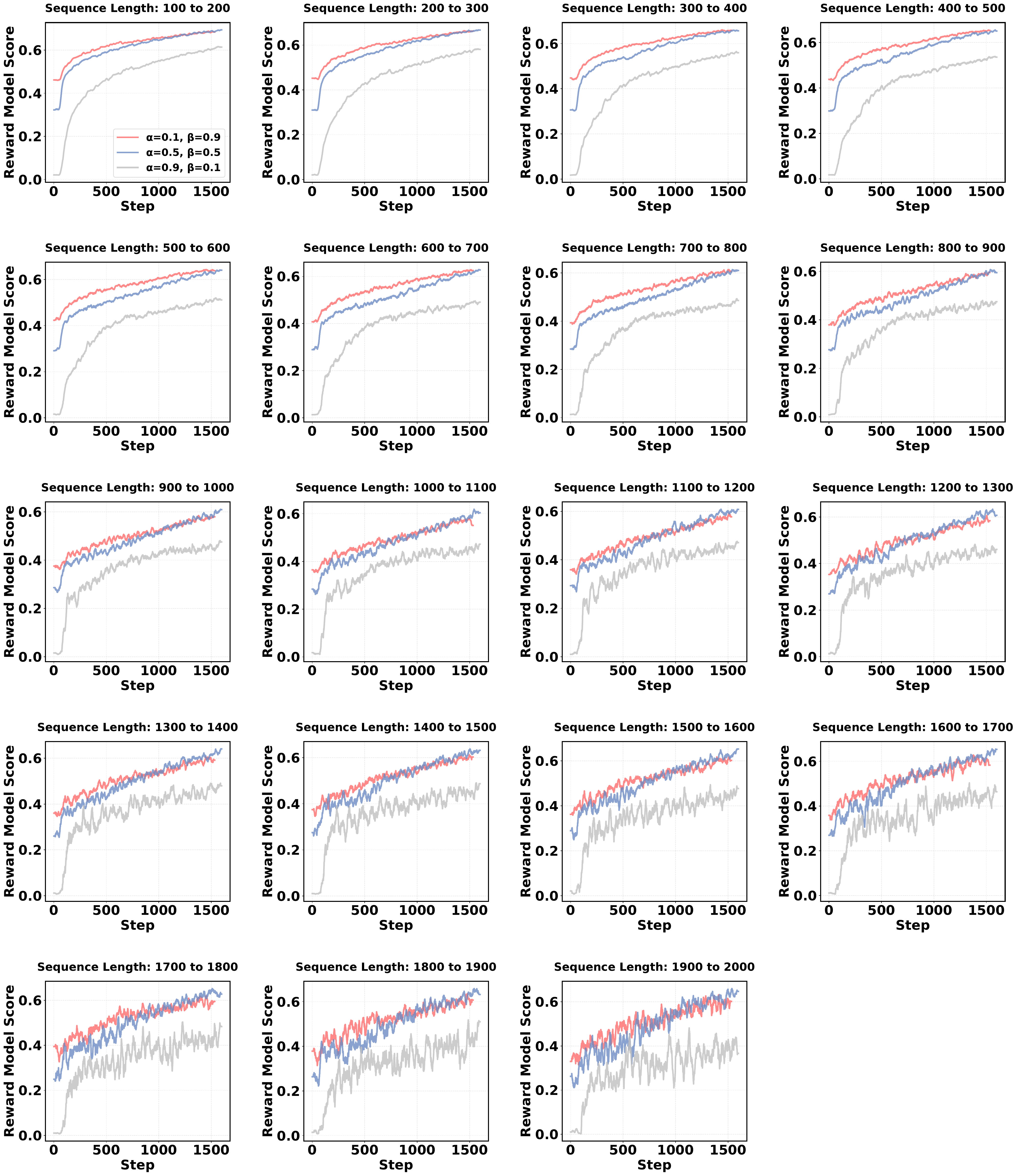}
   \caption{Reward score curve across different BAI ratio. The red curve represents Paradigm 3  BAI ($\alpha=0.1, \beta=0.9$), the blue curve represents Paradigm 3 with BAI ($\alpha=0.5, \beta=0.5$) , while the grey curve represents Paradigm 3 with BAI ($\alpha=0.9, \beta=0.1$). Each figure represents the reward scores of samples with generated lengths in the interval from i×100 to (i+1)×100, where i represents integers ranging from 0 to 19.}
   \label{fig:bai_ratio_comparison}
\end{figure}

To further investigate the influence of different configurations for our BAI approach, we conduct a comprehensive comparison across different merging ratios. Figure~\ref{fig:bai_ratio_comparison} presents the reward score trajectories for three BAI configurations: $(\alpha = 0.1, \beta = 0.9)$ (red curves), $(\alpha = 0.5, \beta = 0.5)$ (blue curves), and $(\alpha = 0.9, \beta = 0.1)$ (grey curves),
analyzed across different sequence length ranges.

The results reveal distinct performance patterns across different sequence length ranges. For shorter sequences, all three configurations demonstrate relatively stable performance, with the balanced $(\alpha = 0.5, \beta = 0.5)$ configuration showing slight advantages in convergence speed and final reward scores. However, as sequence length increases, the differences become more pronounced.
The SFT-heavy configuration $(\alpha = 0.1, \beta = 0.9)$ consistently achieves the highest reward scores across longer sequence ranges, demonstrating superior performance for complex reasoning tasks that require extended chain-of-thought generation. This advantage becomes particularly evident in the long sequence ranges, where the red curves consistently outperform other configurations by substantial margins.
In contrast, the pretrain-heavy configuration $(\alpha = 0.9, \beta = 0.1)$ shows more conservative performance gains, with grey curves typically plateauing at lower reward levels. While this configuration provides stability, it appears to sacrifice some reasoning capability for robustness, particularly in longer sequence generation tasks.
The balanced $(\alpha = 0.5, \beta = 0.5)$ configuration achieves a middle ground, demonstrating competitive performance across all sequence lengths while maintaining training stability. This configuration represents an optimal choice when seeking to balance reasoning capability with training robustness, confirming our earlier analysis of the effectiveness of balanced merging ratios.

\subsection{Addressing Sequence Length Collapse through BAI}

\begin{figure}[htbp]
    \centering
    \begin{subfigure}[b]{0.49\textwidth}
        \centering
        \includegraphics[width=\textwidth]{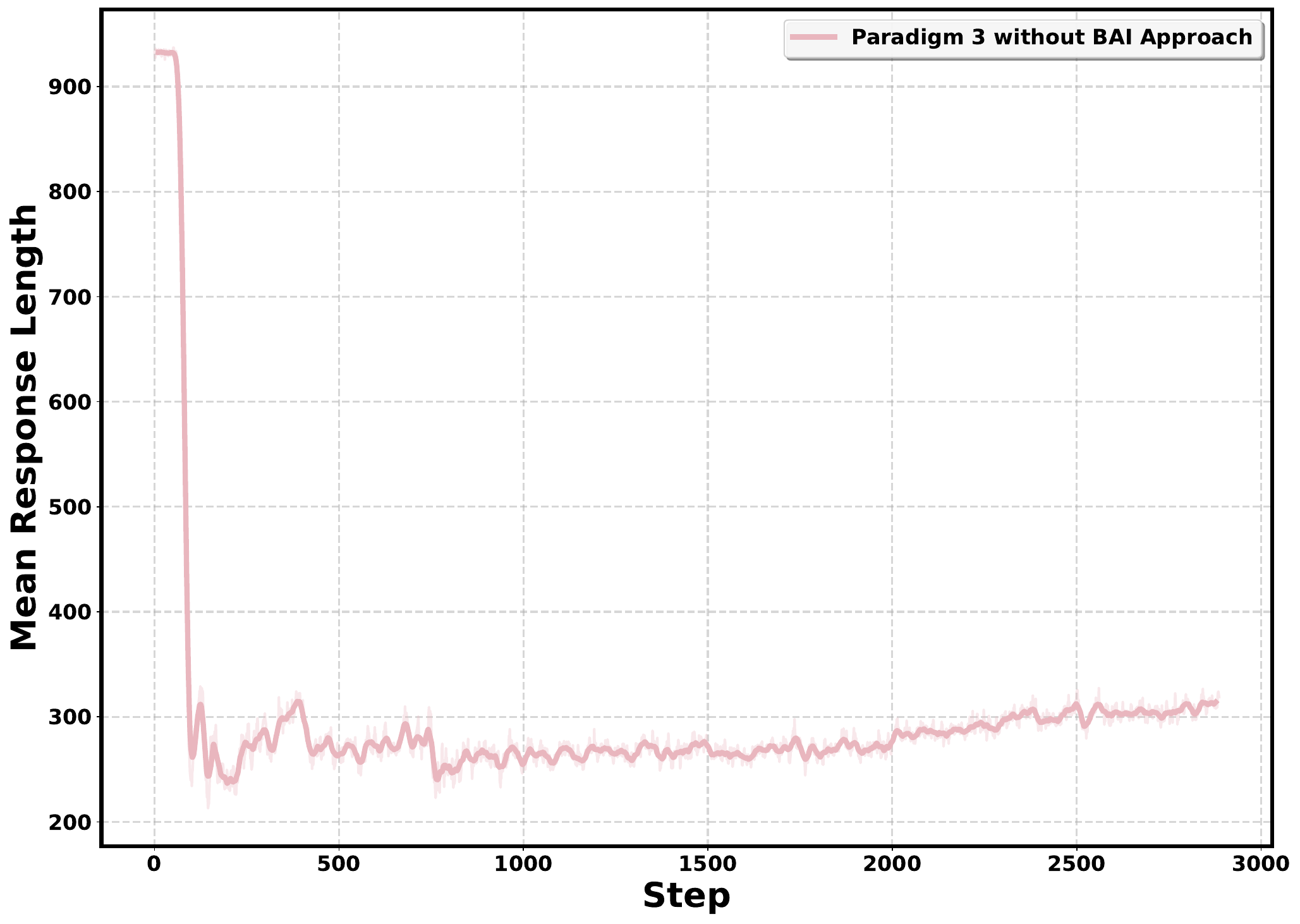}
        \caption{Mean sequence length on Paradigm 3 without BAI Approach.}
        \label{fig:seq_len_baseline}
    \end{subfigure}
    \hfill
    \begin{subfigure}[b]{0.49\textwidth}
        \centering
        \includegraphics[width=\textwidth]{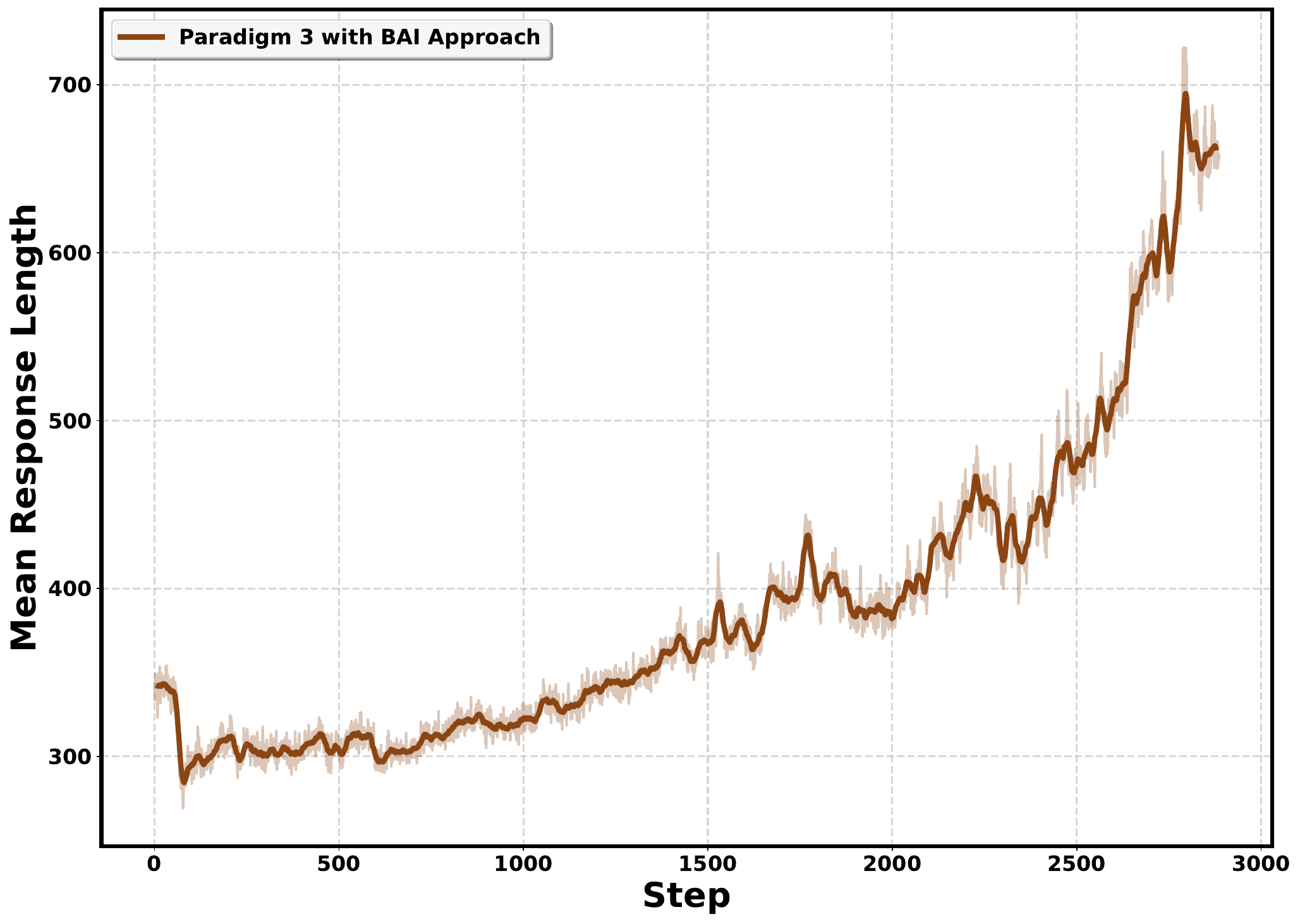}
        \caption{Mean sequence length on BAI ($\alpha=0.6$ $\beta=0.4$ Merging Ratio).}
        \label{fig:seq_len_ratios_more}
    \end{subfigure}
    \caption{3000 steps training: Mean sequence length on Paradigm $3$ with BAI and without BAI.}
\end{figure}

The paradigm $3$ faces another critical challenge in the form of Sequence Length Collapse, where models experience dramatic reduction in sample generation during early RL training phases. As illustrated in Figure~\ref{fig:seq_len_baseline}, models initialized from pure distillation-based reasoning fine-tuned models exhibit catastrophic drops in generated sequence length within the first few training steps, with no recovery throughout the entire training process. This phenomenon fundamentally undermines the model's ability to generate comprehensive reasoning chains and detailed responses.

The sequence length collapse occurs due to the distributional mismatch between specialized reasoning patterns learned during distillation-based fine-tuning and RL optimization requirements. When transitioning to RL training, the reward model optimization creates tension between the specialized SFT patterns and reward signal expectations. This mismatch triggers an over-correction mechanism where the model rapidly shortens responses to achieve higher immediate rewards, effectively engaging in reward hacking by producing concise responses that score well but lack reasoning depth. The concentrated parameter updates during distillation-based reasoning fine-tuning create brittle optimization landscapes that are susceptible to rapid degradation under RL gradient updates.

\begin{figure}[H]
\centering
\begin{tabular}{cc}
\includegraphics[width=0.4\textwidth]{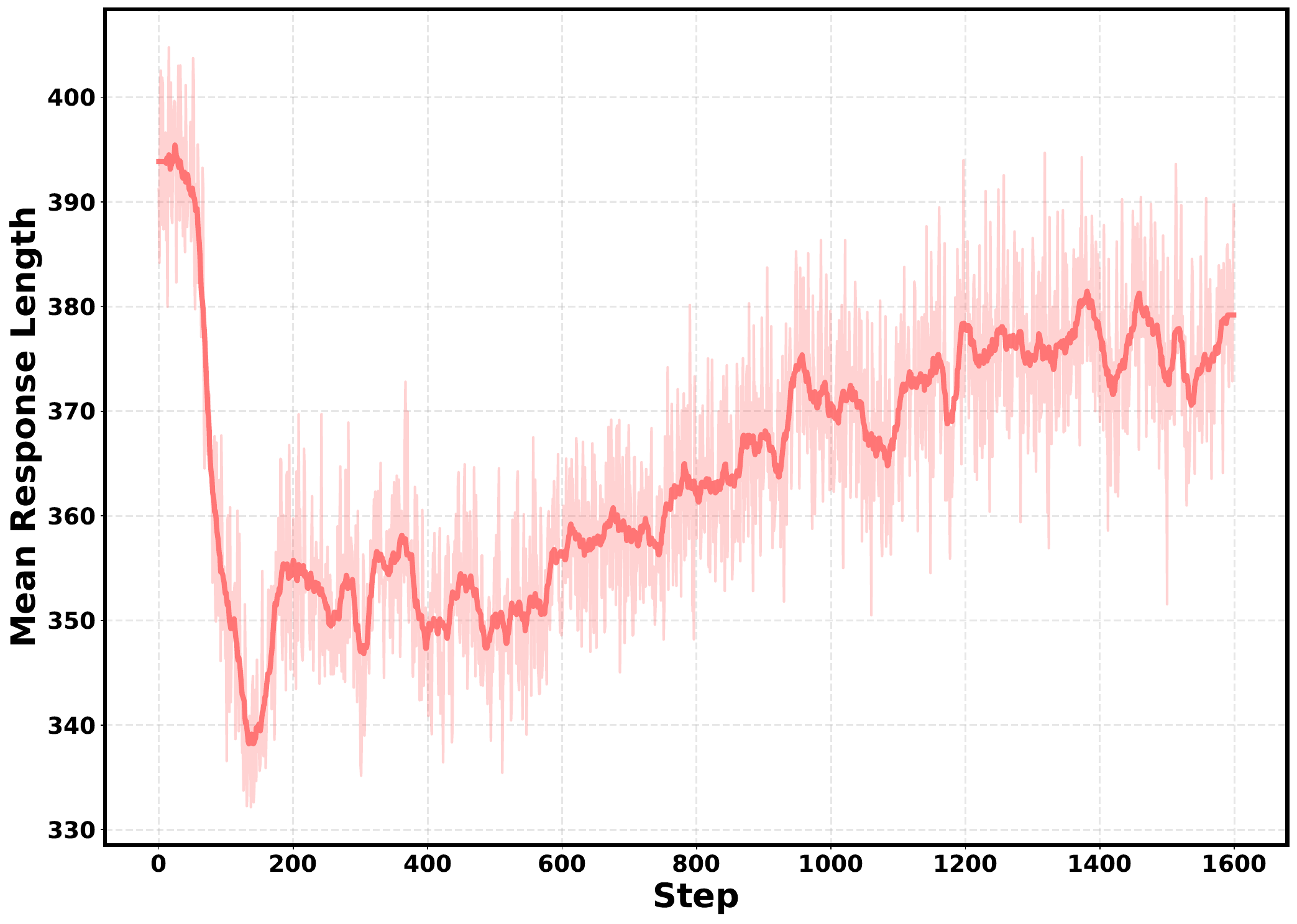} &
\includegraphics[width=0.4\textwidth]{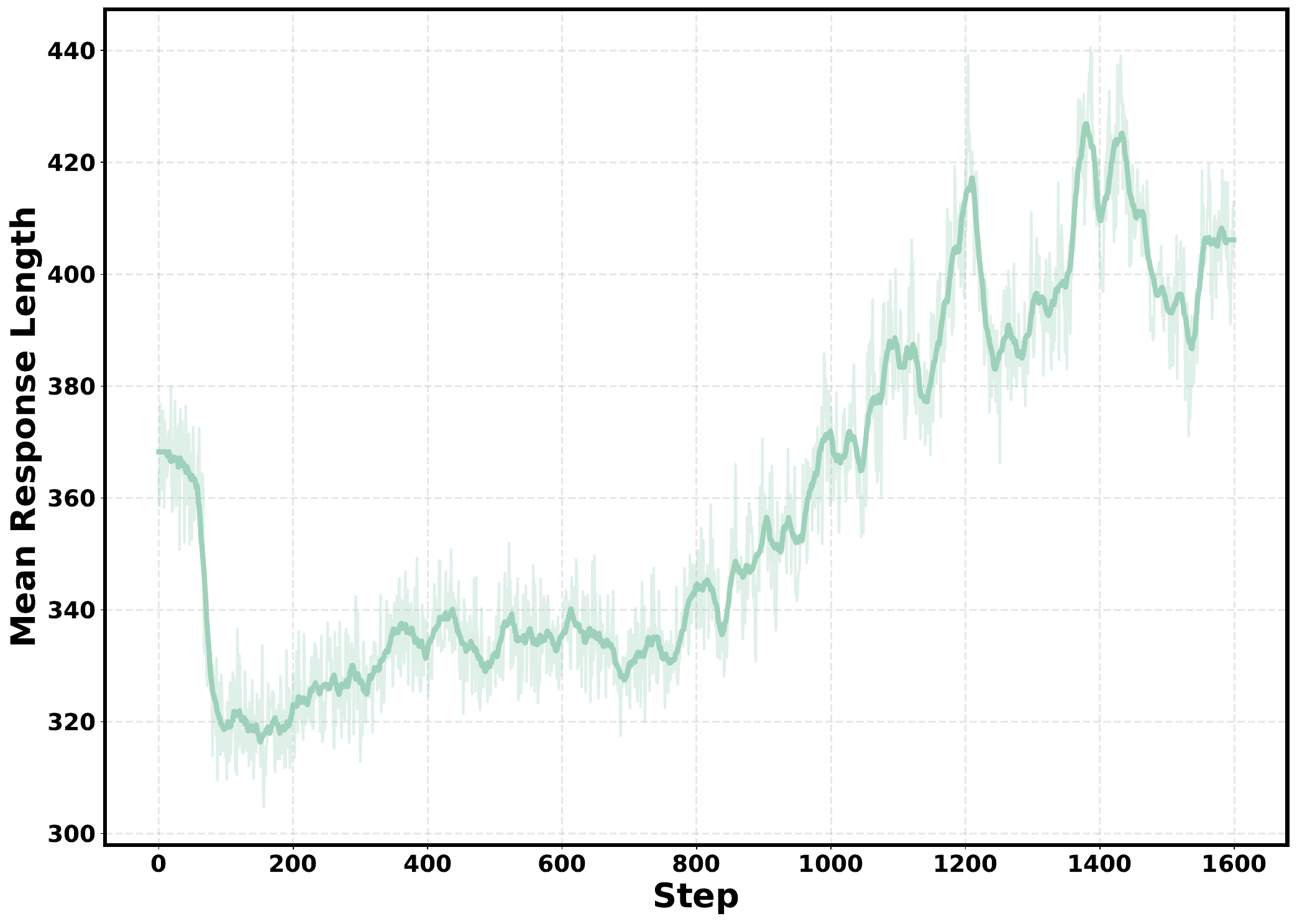} \\
BAI: $\alpha=0.1, \beta=0.9$ & BAI: $\alpha=0.4, \beta=0.6$ \\[0.3em]
\includegraphics[width=0.4\textwidth]{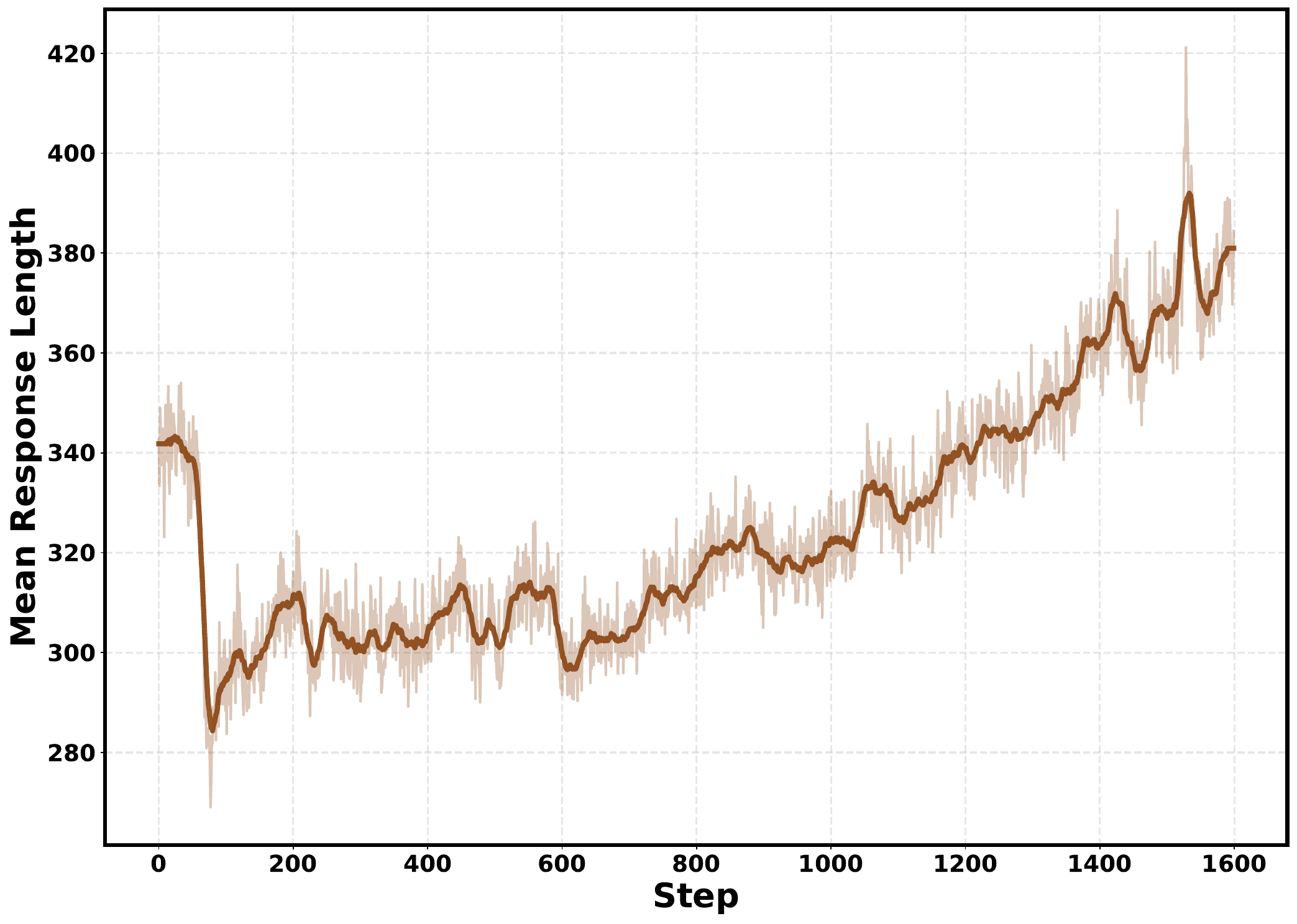} &
\includegraphics[width=0.4\textwidth]{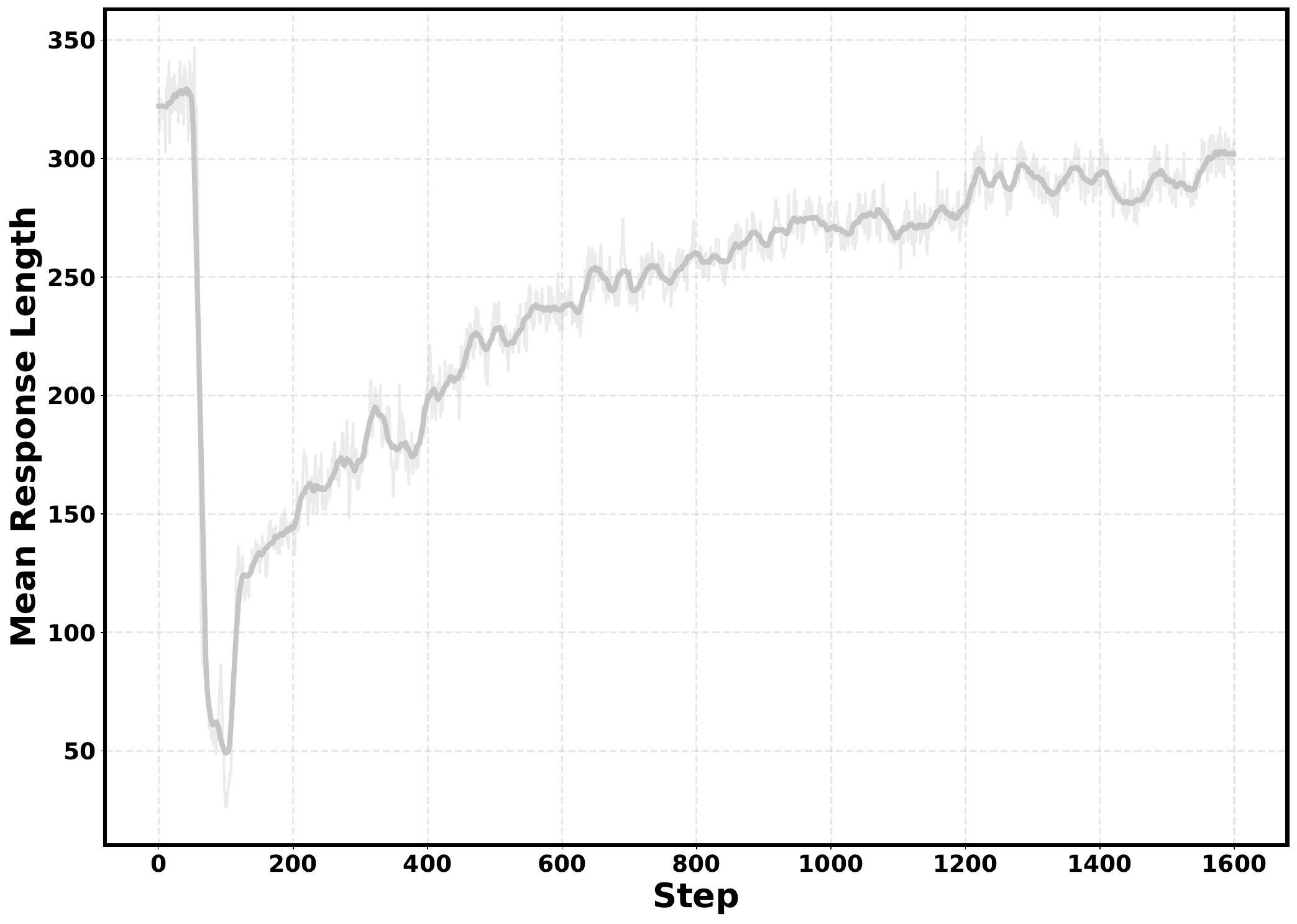} \\
BAI: $\alpha=0.6, \beta=0.4$ & BAI: $\alpha=0.9, \beta=0.1$ \\
\end{tabular}
\caption{Mean sequence length across different BAI merging ratios. More figures across different ratios can be found in the appendix~\ref{sec:mean_response_len}.}
\label{fig:seq_len_part}
\end{figure}

\begin{figure}[htbp]
\centering
\begin{tabular}{ccc}
\includegraphics[width=0.33\textwidth]{images/seq_len/12.5_seq_0.1.pdf} &
\includegraphics[width=0.33\textwidth]{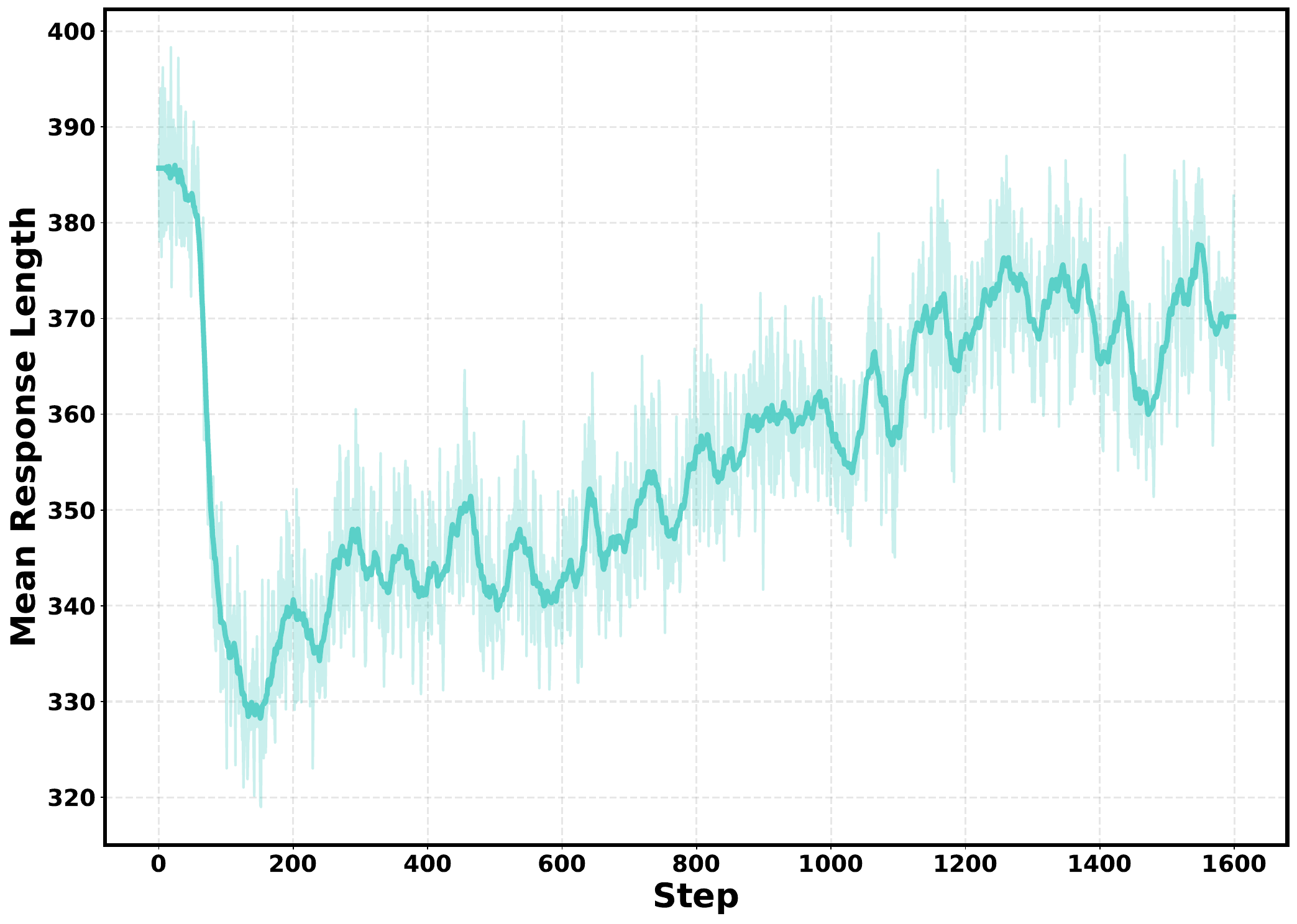} &
\includegraphics[width=0.33\textwidth]{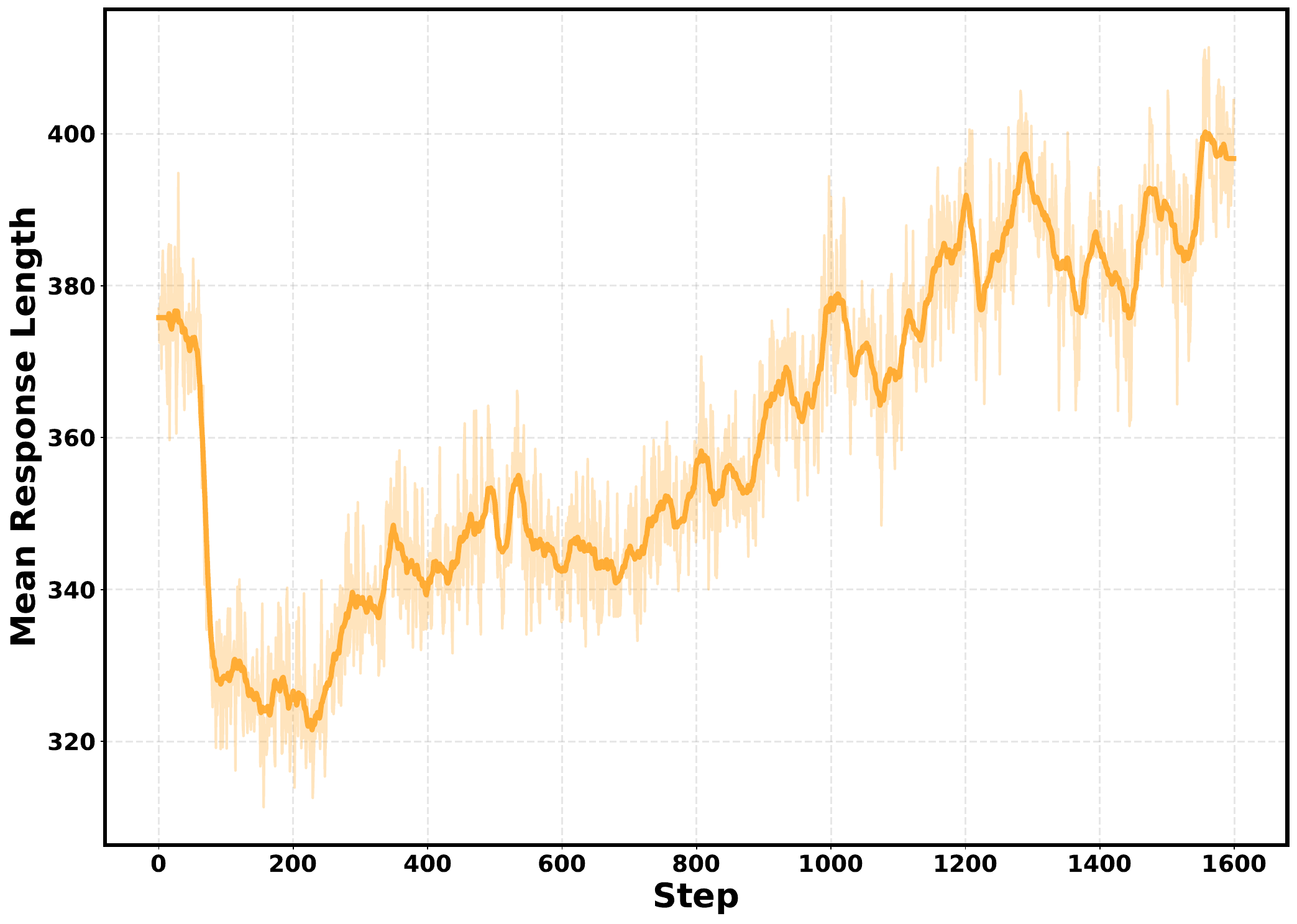} \\
BAI: $\alpha=0.1, \beta=0.9$ & BAI: $\alpha=0.2, \beta=0.8$ & BAI: $\alpha=0.3, \beta=0.7$ \\[0.5em]
\includegraphics[width=0.33\textwidth]{images/seq_len/12.5_seq_0.4.pdf} &
\includegraphics[width=0.33\textwidth]{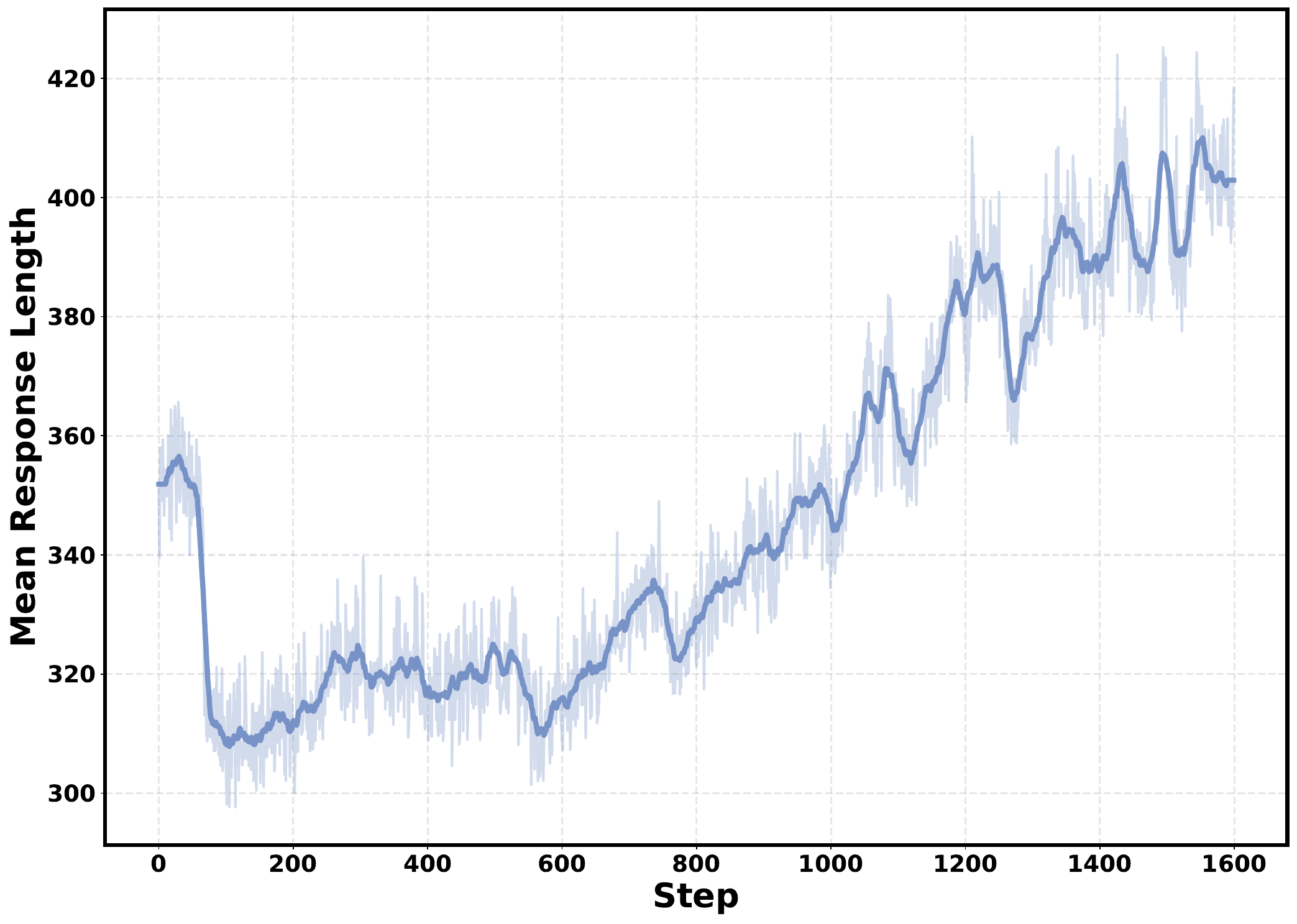} &
\includegraphics[width=0.33\textwidth]{images/seq_len/12.5_seq_0.6.pdf} \\
BAI: $\alpha=0.4, \beta=0.6$ & BAI: $\alpha=0.5, \beta=0.5$ & BAI: $\alpha=0.6, \beta=0.4$ \\[0.5em]
\includegraphics[width=0.33\textwidth]{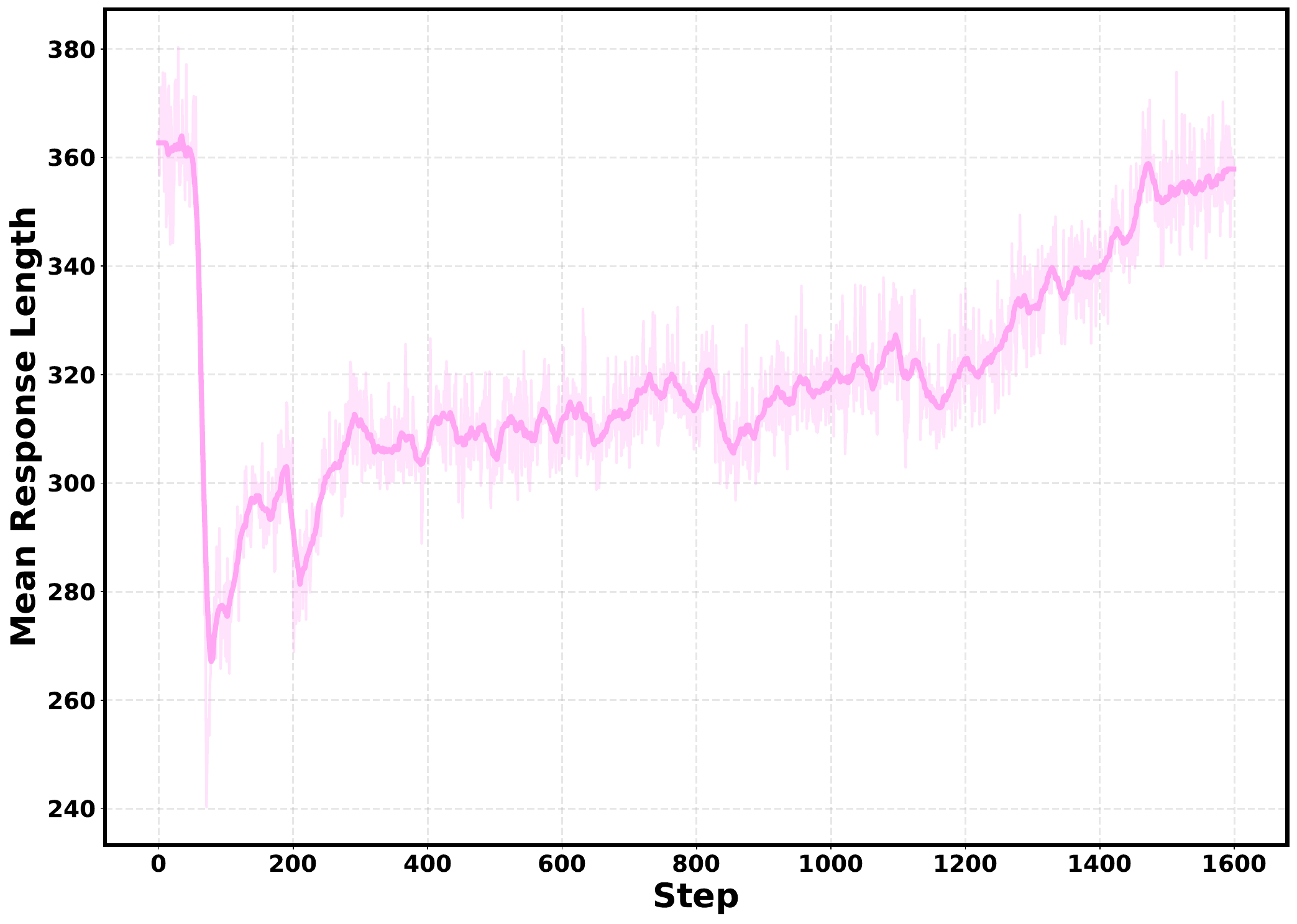} &
\includegraphics[width=0.33\textwidth]{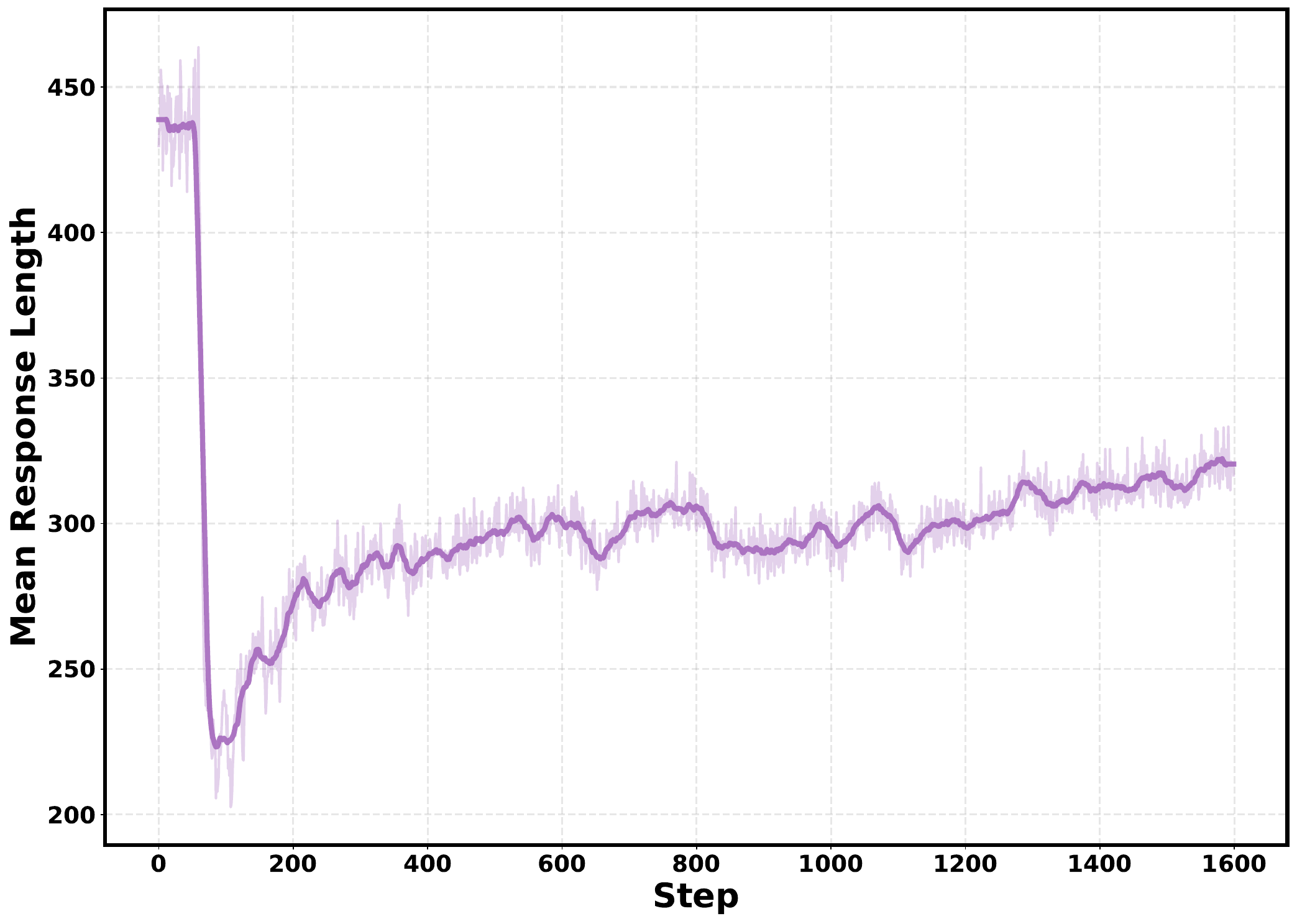} &
\includegraphics[width=0.33\textwidth]{images/seq_len/12.5_seq_0.9.pdf} \\
BAI: $\alpha=0.7, \beta=0.3$ & BAI: $\alpha=0.8, \beta=0.2$ & BAI: $\alpha=0.9, \beta=0.1$ \\
\end{tabular}
\caption{Mean sequence length on different BAI merging ratio.}
\label{fig:seq_len_full}
\end{figure}

To address this challenge, we evaluate our BAI approach across different merging ratios, examining how the balance between pretrained and SFT parameters affects sequence length stability. Figure~\ref{fig:seq_len_part} and Figure~\ref{fig:seq_len_full} present the mean sequence lengths for different merging ratios during RL training. The results reveal consistent behavior: every merging ratio effectively reduces or eliminates the initial sequence length collapse compared to the pure SFT baseline. More importantly, ratios closer to the balanced $(\alpha = 0.5, \beta = 0.5)$ configuration exhibit the most desirable behavior—not only do they prevent the initial collapse, but they also demonstrate progressive sequence length growth throughout training.

This progressive improvement in balanced merging ratios can be attributed to the optimal equilibrium between knowledge preservation and reasoning capability. Ratios heavily weighted toward the pretrained model (e.g., $(\alpha = 0.9, \beta = 0.1)$) lack sufficient reasoning initialization, requiring longer training to develop CoT capabilities. Conversely, ratios favoring the SFT model (e.g., $(\alpha = 0.1, \beta = 0.9)$) retain more reasoning patterns but inherit instability from the specialized fine-tuning. The balanced $(\alpha = 0.5, \beta = 0.5)$ and $(\alpha = 0.6, \beta = 0.4)$ ratios achieve an optimal compromise, preserving enough pretrained stability to prevent collapse while maintaining sufficient reasoning capability to enable progressive development.

Importantly, the sequence length growth observed in balanced ratios highlights that optimal model development requires consideration of both performance metrics and training dynamics. While SFT-heavy configurations may achieve higher immediate benchmark scores, the progressive sequence length improvement in balanced ratios demonstrates healthier learning patterns that are more likely to sustain long-term capability development. This finding underscores that effective reasoning model training should not solely pursue metric optimization but must also ensure stable and progressive training states.

To further validate the long-term sequence length growth of our approach, we extended training for the $(\alpha = 0.6, \beta = 0.4)$ configuration by an additional $1,400$ steps. As illustrated in Figure~\ref{fig:seq_len_ratios_more}, sequence length continues to grow progressively with training steps. This demonstrates that our BAI approach not only prevents initial collapse but also establishes a foundation for sustained capability development, suggesting that longer training could yield substantial improvements in reasoning depth and comprehensiveness.

This balance creates a more robust optimization landscape that supports both immediate stability and long-term capability growth, establishing the foundation for effective reasoning model development in the third paradigm.

\subsection{KL Divergence Analysis of BAI Configurations}

To better understand the training dynamics enabled by our BAI approach, we analyze the KL divergence between the training policy and the sampling policy throughout the RL training process. The KL divergence serves as a critical indicator of policy stability during optimization, with effective training typically characterized by controlled divergence patterns that avoid both excessive drift and stagnation.

\begin{figure}[htbp]
  \centering
  \includegraphics[width=0.7\textwidth]{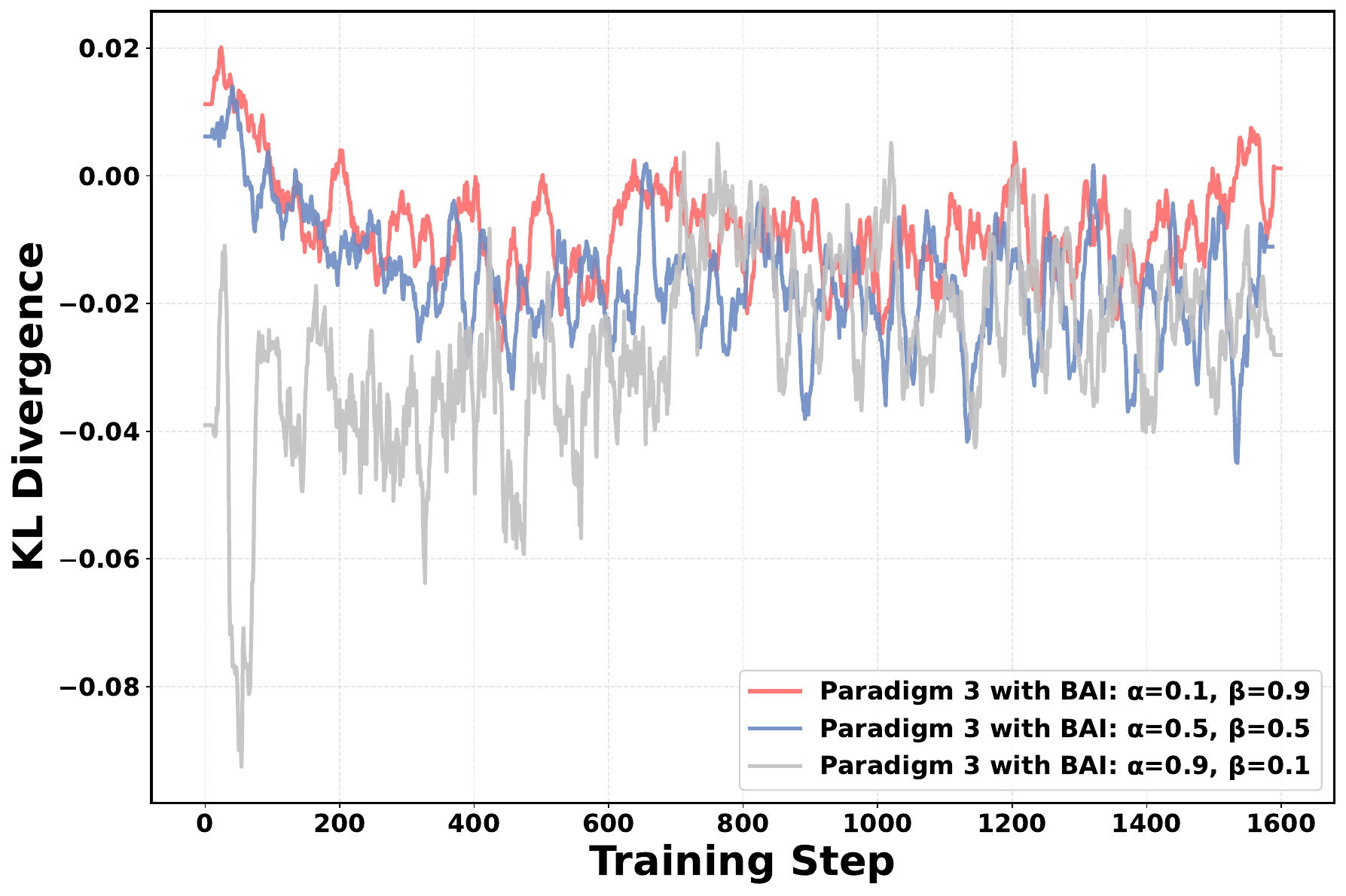}
  \caption{KL divergence analysis across BAI configurations.}
  \label{fig:kl_divergence}
\end{figure}

Figure~\ref{fig:kl_divergence} presents the KL divergence trajectories for three representative BAI configurations: $(\alpha = 0.1, \beta = 0.9)$ (SFT-heavy), $(\alpha = 0.5, \beta = 0.5)$ (balanced), and $(\alpha = 0.9, \beta = 0.1)$ (pre-train heavy). The results reveal distinct behavioral patterns that correlate with the merging characteristics and provide insight into the underlying optimization dynamics.
The BAI $(\alpha = 0.1, \beta = 0.9)$ configuration demonstrates the most stable KL divergence patterns with minimal fluctuations throughout training. This stability stems from the specialized reasoning patterns acquired during fine-tuning, which provide consistent policy behavior under RL optimization.

In contrast, the $(\alpha = 0.9, \beta = 0.1)$ configuration exhibits the most volatile KL divergence pattern with frequent spikes and irregular fluctuations. This instability reflects the challenge of adapting the heavy pretrained parameters to specific RL objectives, where the broad parameter configuration struggles to maintain consistent policy behavior.

The balanced BAI $(\alpha = 0.5, \beta = 0.5)$ configuration achieves a great ground, exhibiting moderate KL divergence fluctuations that indicate healthy optimization dynamics. While showing more variation than the SFT-heavy configuration, it maintains better stability than the pretrain-heavy setup, demonstrating effective adaptation while preserving training consistency.

These KL divergence patterns provide additional evidence supporting our BAI approach and help explain the superior performance of balanced configurations. The controlled divergence in balanced merging ratios indicates that BAI enables effective adaptation to reward signals while maintaining training stability, establishing favorable conditions for sustained learning and capability growth.

\section{Future Directions}

Although our BAI approach demonstrates significant effectiveness in addressing training instabilities in the third paradigm, several directions warrant further investigation. Future work could explore adaptive merging strategies that dynamically adjust weights during training, investigate the application of similar approaches to other model architectures and training objectives to better understand the optimization dynamics underlying these phenomena. Additionally, extending our analysis to other specialized fine-tuning domains beyond reasoning, such as agent training and multi-modalities, could validate the broader applicability of weighted model merging strategies.

The success of BAI establishes that proper actor initialization is fundamental to stable training in the third paradigm. By demonstrating that strategic model weight interpolation can create robust initializations that prevent training instabilities, this work highlights the critical importance of initialization strategies in modern language model development. Our findings open new avenues for developing initialization methodologies that enable reasoning models to successfully leverage both distillation efficiency and human feedback optimization while maintaining training stability throughout the process.

\section{Conclusion}

In this paper, we investigate the development of reasoning capabilities in large language models through three established paradigms: the instruction tuning and alignment paradigm (Paradigm 1), distillation-based reasoning fine-tuning (Paradigm 2), and their combination through applying RLHF to distillation-trained models (Paradigm 3). Although Paradigms 1 and 2 have proven effective independently, Paradigm 3 faces critical training instabilities.
Our analysis identified two fundamental challenges in Paradigm 3: Sequence Length Collapse, where models experience dramatic reduction in language generation during early RL training, and the Reward Hockey Stick Curve, featuring initial reward score degradation followed by gradual recovery. These phenomena fundamentally compromise the model's ability to maintain detailed reasoning chains and stable training dynamics.
To address these challenges, we proposed Balanced Actor Initialization (BAI), a two-stage weighted model merging approach. BAI creates robust actor initializations that prevent training instabilities while maintaining specialized reasoning capabilities.
Our experimental evaluation across diverse benchmarks demonstrates BAI's effectiveness. BAI outperforms all three individual paradigms while maintaining stable training throughout the RL process. The approach consistently eliminates sequence length collapse, mitigates reward curve instabilities, and enables continuous sequence length improvement during training. These results confirm that BAI successfully enables stable training in Paradigm 3, allowing practitioners to leverage both the efficiency of distillation and the alignment benefits of reinforcement learning from human feedback optimization.

\newpage

\section*{Contributions}
\label{contributions}
\textbf{Project Lead}\quad 

Chen Zheng


\textbf{Algorithm}

Chen Zheng, Yiyuan Ma, Yuan Yang, Deyi Liu, Jing Liu


\textbf{Infrastructure}

Zuquan Song, Yuxin Song, Cheng Ren, Hang Zhu, Xin Liu



\textbf{Supervision}

Yiyuan Ma, Siyuan Qiao, Xun Zhou, Liang Xiang, Yonghui Wu


\textbf{Affiliation}

ByteDance Seed

\section*{Acknowledgements}
We thank Yuhang Cai, Yunshui Li, Jin Ma, Chaoyi Zhang, Yutao Zeng, Jianqiao Lu, Minrui Wang, Shiyi Zhan, Jie Huang, Yao Luo, Xu Ouyang, Chenglin Yang, Xia Xiao, Shen Zheng, Ran Xin, Zaiyuan Wang, Mengyao Zhang, Lin Yan, Jiecao Chen, Kai Shen, and other members at Bytedance for the support for this paper.

\clearpage

\bibliographystyle{plainnat}
\bibliography{main}

\begin{thebibliography}{43}
\providecommand{\natexlab}[1]{#1}
\providecommand{\url}[1]{\texttt{#1}}
\expandafter\ifx\csname urlstyle\endcsname\relax
  \providecommand{\doi}[1]{doi: #1}\else
  \providecommand{\doi}{doi: \begingroup \urlstyle{rm}\Url}\fi

\bibitem[Achiam et~al.(2023)Achiam, Adler, Agarwal, Ahmad, Akkaya, Aleman, Almeida, Altenschmidt, Altman, Anadkat, et~al.]{achiam2023gpt}
Josh Achiam, Steven Adler, Sandhini Agarwal, Lama Ahmad, Ilge Akkaya, Florencia~Leoni Aleman, Diogo Almeida, Janko Altenschmidt, Sam Altman, Shyamal Anadkat, et~al.
\newblock Gpt-4 technical report.
\newblock \emph{arXiv preprint arXiv:2303.08774}, 2023.

\bibitem[Akiba et~al.(2024)Akiba, Sano, Yanai, Ohta, and Koyama]{akiba2024evolutionary}
Takuya Akiba, Makoto Sano, Toshihiko Yanai, Kengo Ohta, and Masanori Koyama.
\newblock Evolutionary optimization of model merging recipes.
\newblock \emph{arXiv preprint arXiv:2403.13187}, 2024.

\bibitem[Austin et~al.(2021)Austin, Odena, Nye, Bosma, Michalewski, Dohan, Jiang, Cai, Terry, Le, et~al.]{austin2021program}
Jacob Austin, Augustus Odena, Maxwell Nye, Maarten Bosma, Henryk Michalewski, David Dohan, Ellen Jiang, Carrie Cai, Michael Terry, Quoc Le, et~al.
\newblock Program synthesis with large language models.
\newblock \emph{arXiv preprint arXiv:2108.07732}, 2021.

\bibitem[Bai et~al.(2022{\natexlab{a}})Bai, Jones, Ndousse, Askell, Chen, DasSarma, Drain, Fort, Ganguli, Henighan, et~al.]{bai2022training}
Yuntao Bai, Andy Jones, Kamal Ndousse, Amanda Askell, Anna Chen, Nova DasSarma, Dawn Drain, Stanislav Fort, Deep Ganguli, Tom Henighan, et~al.
\newblock Training a helpful and harmless assistant with reinforcement learning from human feedback.
\newblock \emph{arXiv preprint arXiv:2204.05862}, 2022{\natexlab{a}}.

\bibitem[Bai et~al.(2022{\natexlab{b}})Bai, Kadavath, Kundu, Askell, Kernion, Jones, Chen, Goldie, Mirhoseini, McKinnon, et~al.]{bai2022constitutional}
Yuntao Bai, Saurav Kadavath, Sandipan Kundu, Amanda Askell, Jackson Kernion, Andy Jones, Anna Chen, Anna Goldie, Azalia Mirhoseini, Cameron McKinnon, et~al.
\newblock Constitutional ai: Harmlessness from ai feedback.
\newblock \emph{arXiv preprint arXiv:2212.08073}, 2022{\natexlab{b}}.

\bibitem[Chen et~al.(2025)Chen, Qin, Liu, Peng, Guan, Wang, Hu, Zhou, Gao, and Che]{chen2025towards}
Qiguang Chen, Libo Qin, Jinhao Liu, Dengyun Peng, Jiannan Guan, Peng Wang, Mengkang Hu, Yuhang Zhou, Te~Gao, and Wanxiang Che.
\newblock Towards reasoning era: A survey of long chain-of-thought for reasoning large language models.
\newblock \emph{arXiv preprint arXiv:2503.09567}, 2025.

\bibitem[Dao et~al.(2022)Dao, Fu, Ermon, Rudra, and R{\'e}]{dao2022flashattention}
Tri Dao, Dan Fu, Stefano Ermon, Atri Rudra, and Christopher R{\'e}.
\newblock Flashattention: Fast and memory-efficient exact attention with io-awareness.
\newblock \emph{Advances in neural information processing systems}, 35:\penalty0 16344--16359, 2022.

\bibitem[DeepSeek-AI et~al.(2024)]{deepseek2024}
DeepSeek-AI et~al.
\newblock Deepseek-r1: Incentivizing reasoning capability with reinforcement learning.
\newblock \emph{arXiv preprint arXiv:2501.12948}, 2024.

\bibitem[Du et~al.(2025)Du, Yao, Ma, Wang, Zheng, Zhu, Liu, Liang, Jin, Wei, et~al.]{du2025supergpqa}
Xinrun Du, Yifan Yao, Kaijing Ma, Bingli Wang, Tianyu Zheng, King Zhu, Minghao Liu, Yiming Liang, Xiaolong Jin, Zhenlin Wei, et~al.
\newblock Supergpqa: Scaling llm evaluation across 285 graduate disciplines.
\newblock \emph{arXiv preprint arXiv:2502.14739}, 2025.

\bibitem[Goddard et~al.(2024)Goddard, Siriwardhana, Ehghaghi, Meyers, Karpukhin, Benedict, McQuade, and Solawetz]{goddard2024arcee}
Charles Goddard, Shamane Siriwardhana, Malikeh Ehghaghi, Luke Meyers, Vlad Karpukhin, Brian Benedict, Mark McQuade, and Jacob Solawetz.
\newblock Arcee's mergekit: A toolkit for merging large language models.
\newblock \emph{arXiv preprint arXiv:2403.13257}, 2024.

\bibitem[Guo et~al.(2025)Guo, Yang, Zhang, Song, Zhang, Xu, Zhu, Ma, Wang, Bi, et~al.]{guo2025deepseek}
Daya Guo, Dejian Yang, Haowei Zhang, Junxiao Song, Ruoyu Zhang, Runxin Xu, Qihao Zhu, Shirong Ma, Peiyi Wang, Xiao Bi, et~al.
\newblock Deepseek-r1: Incentivizing reasoning capability in llms via reinforcement learning.
\newblock \emph{arXiv preprint arXiv:2501.12948}, 2025.

\bibitem[Hendrycks et~al.(2020)Hendrycks, Burns, Basart, Zou, Mazeika, Song, and Steinhardt]{hendrycks2020measuring}
Dan Hendrycks, Collin Burns, Steven Basart, Andy Zou, Mantas Mazeika, Dawn Song, and Jacob Steinhardt.
\newblock Measuring massive multitask language understanding.
\newblock \emph{arXiv preprint arXiv:2009.03300}, 2020.

\bibitem[Hendrycks et~al.(2021)Hendrycks, Burns, Kadavath, Arora, Basart, Tang, Song, and Steinhardt]{hendrycks2021measuring}
Dan Hendrycks, Collin Burns, Saurav Kadavath, Akul Arora, Steven Basart, Eric Tang, Dawn Song, and Jacob Steinhardt.
\newblock Measuring mathematical problem solving with the math dataset.
\newblock \emph{arXiv preprint arXiv:2103.03874}, 2021.

\bibitem[Kim et~al.(2023)Kim, Park, Kim, Lee, Song, Kim, Kim, Kim, Lee, Kim, et~al.]{kim2023solar}
Dahyun Kim, Chanjun Park, Sanghoon Kim, Wonsung Lee, Wonho Song, Yunsu Kim, Hyeonwoo Kim, Yungi Kim, Hyeonju Lee, Jihoo Kim, et~al.
\newblock Solar 10.7 b: Scaling large language models with simple yet effective depth up-scaling.
\newblock \emph{arXiv preprint arXiv:2312.15166}, 2023.

\bibitem[Li et~al.(2025{\natexlab{a}})Li, Cao, Griggs, Liu, Mo, Tang, Hegde, Hakhamaneshi, Patil, Zaharia, et~al.]{li2025llms}
Dacheng Li, Shiyi Cao, Tyler Griggs, Shu Liu, Xiangxi Mo, Eric Tang, Sumanth Hegde, Kourosh Hakhamaneshi, Shishir~G Patil, Matei Zaharia, et~al.
\newblock Llms can easily learn to reason from demonstrations structure, not content, is what matters!
\newblock \emph{arXiv preprint arXiv:2502.07374}, 2025{\natexlab{a}}.

\bibitem[Li et~al.(2024)Li, Chiang, Frick, Dunlap, Wu, Zhu, Gonzalez, and Stoica]{li2024crowdsourced}
Tianle Li, Wei-Lin Chiang, Evan Frick, Lisa Dunlap, Tianhao Wu, Banghua Zhu, Joseph~E Gonzalez, and Ion Stoica.
\newblock From crowdsourced data to high-quality benchmarks: Arena-hard and benchbuilder pipeline.
\newblock \emph{arXiv preprint arXiv:2406.11939}, 2024.

\bibitem[Li et~al.(2025{\natexlab{b}})Li, Ma, Yan, Zhang, Liu, Lu, Xu, Chen, Wang, Zhan, et~al.]{li2025model}
Yunshui Li, Yiyuan Ma, Shen Yan, Chaoyi Zhang, Jing Liu, Jianqiao Lu, Ziwen Xu, Mengzhao Chen, Minrui Wang, Shiyi Zhan, et~al.
\newblock Model merging in pre-training of large language models.
\newblock \emph{arXiv preprint arXiv:2505.12082}, 2025{\natexlab{b}}.

\bibitem[Liu et~al.(2025)Liu, Fu, Ding, Ning, Zhang, Liu, and Zhang]{liu2025logical}
Hanmeng Liu, Zhizhang Fu, Mengru Ding, Ruoxi Ning, Chaoli Zhang, Xiaozhang Liu, and Yue Zhang.
\newblock Logical reasoning in large language models: A survey.
\newblock \emph{arXiv preprint arXiv:2502.09100}, 2025.

\bibitem[Luo et~al.(2025)Luo, Song, Zhang, Liu, Wang, Chen, Su, and Zheng]{luo2025deconstructing}
Yijia Luo, Yulin Song, Xingyao Zhang, Jiaheng Liu, Weixun Wang, GengRu Chen, Wenbo Su, and Bo~Zheng.
\newblock Deconstructing long chain-of-thought: A structured reasoning optimization framework for long cot distillation.
\newblock \emph{arXiv preprint arXiv:2503.16385}, 2025.

\bibitem[Matena and Raffel(2022)]{matena2022merging}
Michael~S Matena and Colin Raffel.
\newblock Merging models with fisher-weighted averaging.
\newblock \emph{Advances in Neural Information Processing Systems}, 35:\penalty0 17703--17716, 2022.

\bibitem[{Mathematical Association of America}(2024)]{aime2024}
{Mathematical Association of America}.
\newblock American invitational mathematics examination 2024.
\newblock \url{https://www.maa.org/math-competitions/american-invitational-mathematics-examination-aime}, 2024.

\bibitem[Matsuoka(2022)]{matsuoka2022merging}
Takuya Matsuoka.
\newblock Merging distributed neural networks with bayesian learning.
\newblock \emph{arXiv preprint arXiv:2204.06132}, 2022.

\bibitem[Ni et~al.(2024)Ni, Xue, Yue, Deng, Shah, Jain, Neubig, and You]{ni2024mixeval}
Jinjie Ni, Fuzhao Xue, Xiang Yue, Yuntian Deng, Mahir Shah, Kabir Jain, Graham Neubig, and Yang You.
\newblock Mixeval: Deriving wisdom of the crowd from llm benchmark mixtures.
\newblock \emph{Advances in Neural Information Processing Systems}, 37:\penalty0 98180--98212, 2024.

\bibitem[OpenAI(2024)]{openai2024o1}
OpenAI.
\newblock Learning to reason with llms.
\newblock \url{https://openai.com/index/learning-to-reason-with-llms/}, 2024.

\bibitem[Ouyang et~al.(2022)Ouyang, Wu, Jiang, Almeida, Wainwright, Mishkin, Zhang, Agarwal, Slama, Ray, et~al.]{ouyang2022training}
Long Ouyang, Jeffrey Wu, Xu~Jiang, Diogo Almeida, Carroll Wainwright, Pamela Mishkin, Chong Zhang, Sandhini Agarwal, Katarina Slama, Alex Ray, et~al.
\newblock Training language models to follow instructions with human feedback.
\newblock \emph{Advances in neural information processing systems}, 35:\penalty0 27730--27744, 2022.

\bibitem[Rafailov et~al.(2024)Rafailov, Sharma, Mitchell, Manning, Ermon, and Finn]{rafailov2024direct}
Rafael Rafailov, Archit Sharma, Eric Mitchell, Christopher~D Manning, Stefano Ermon, and Chelsea Finn.
\newblock Direct preference optimization: Your language model is secretly a reward model.
\newblock \emph{Advances in Neural Information Processing Systems}, 36, 2024.

\bibitem[Schulman et~al.(2017)Schulman, Wolski, Dhariwal, Radford, and Klimov]{schulman2017proximal}
John Schulman, Filip Wolski, Prafulla Dhariwal, Alec Radford, and Oleg Klimov.
\newblock Proximal policy optimization algorithms.
\newblock \emph{arXiv preprint arXiv:1707.06347}, 2017.

\bibitem[Seed et~al.(2025)Seed, Chen, Fan, Liu, Liu, Lin, Wang, Wang, Wei, Xu, et~al.]{seed2025seed1}
ByteDance Seed, Jiaze Chen, Tiantian Fan, Xin Liu, Lingjun Liu, Zhiqi Lin, Mingxuan Wang, Chengyi Wang, Xiangpeng Wei, Wenyuan Xu, et~al.
\newblock Seed1. 5-thinking: Advancing superb reasoning models with reinforcement learning.
\newblock \emph{arXiv preprint arXiv:2504.13914}, 2025.

\bibitem[Shao et~al.(2024)Shao, Wang, Zhu, Xu, Song, Bi, Zhang, Zhang, Li, Wu, et~al.]{shao2024deepseekmath}
Zhihong Shao, Peiyi Wang, Qihao Zhu, Runxin Xu, Junxiao Song, Xiao Bi, Haowei Zhang, Mingchuan Zhang, YK~Li, Y~Wu, et~al.
\newblock Deepseekmath: Pushing the limits of mathematical reasoning in open language models.
\newblock \emph{arXiv preprint arXiv:2402.03300}, 2024.

\bibitem[Shoeybi et~al.(2019)Shoeybi, Patwary, Puri, LeGresley, Casper, and Catanzaro]{shoeybi2019megatron}
Mohammad Shoeybi, Mostofa Patwary, Raul Puri, Patrick LeGresley, Jared Casper, and Bryan Catanzaro.
\newblock Megatron-lm: Training multi-billion parameter language models using model parallelism.
\newblock \emph{arXiv preprint arXiv:1909.08053}, 2019.

\bibitem[Sui et~al.(2025)Sui, Chuang, Wang, Zhang, Zhang, Yuan, Liu, Wen, Zhong, Chen, et~al.]{sui2025stop}
Yang Sui, Yu-Neng Chuang, Guanchu Wang, Jiamu Zhang, Tianyi Zhang, Jiayi Yuan, Hongyi Liu, Andrew Wen, Shaochen Zhong, Hanjie Chen, et~al.
\newblock Stop overthinking: A survey on efficient reasoning for large language models.
\newblock \emph{arXiv preprint arXiv:2503.16419}, 2025.

\bibitem[Tang et~al.(2024)Tang, Tran, Nguyen, Chen, Zhu, Ahmad, et~al.]{tang2024dare}
Yu~Tang, Trung Tran, Linh Nguyen, Ping Chen, Yawen Zhu, Ali Ahmad, et~al.
\newblock Dare: Drop and rescale for model merging.
\newblock \emph{arXiv preprint arXiv:2311.03099}, 2024.

\bibitem[Team et~al.(2023)Team, Anil, Borgeaud, Alayrac, Yu, Soricut, Schalkwyk, Dai, Hauth, Millican, et~al.]{team2023gemini}
Gemini Team, Rohan Anil, Sebastian Borgeaud, Jean-Baptiste Alayrac, Jiahui Yu, Radu Soricut, Johan Schalkwyk, Andrew~M Dai, Anja Hauth, Katie Millican, et~al.
\newblock Gemini: a family of highly capable multimodal models.
\newblock \emph{arXiv preprint arXiv:2312.11805}, 2023.

\bibitem[Touvron et~al.(2023)Touvron, Martin, Stone, Albert, Almahairi, Babaei, Bashlykov, Batra, Bhargava, Bhosale, et~al.]{touvron2023llama}
Hugo Touvron, Louis Martin, Kevin Stone, Peter Albert, Amjad Almahairi, Yasmine Babaei, Nikolay Bashlykov, Soumya Batra, Prajjwal Bhargava, Shruti Bhosale, et~al.
\newblock Llama 2: Open foundation and fine-tuned chat models.
\newblock \emph{arXiv preprint arXiv:2307.09288}, 2023.

\bibitem[Wang et~al.(2024)Wang, Ma, Zhang, Ni, Chandra, Guo, Ren, Arulraj, He, Jiang, et~al.]{wang2024mmlu}
Yubo Wang, Xueguang Ma, Ge~Zhang, Yuansheng Ni, Abhranil Chandra, Shiguang Guo, Weiming Ren, Aaran Arulraj, Xuan He, Ziyan Jiang, et~al.
\newblock Mmlu-pro: A more robust and challenging multi-task language understanding benchmark.
\newblock \emph{Advances in Neural Information Processing Systems}, 37:\penalty0 95266--95290, 2024.

\bibitem[Wei et~al.(2022)Wei, Wang, Schuurmans, Bosma, Xia, Chi, Le, Zhou, et~al.]{wei2022chain}
Jason Wei, Xuezhi Wang, Dale Schuurmans, Maarten Bosma, Fei Xia, Ed~Chi, Quoc~V Le, Denny Zhou, et~al.
\newblock Chain-of-thought prompting elicits reasoning in large language models.
\newblock \emph{Advances in neural information processing systems}, 35:\penalty0 24824--24837, 2022.

\bibitem[White et~al.(2024)White, Dooley, Roberts, Pal, Feuer, Jain, Shwartz-Ziv, Jain, Saifullah, Naidu, et~al.]{white2024livebench}
Colin White, Samuel Dooley, Manley Roberts, Arka Pal, Ben Feuer, Siddhartha Jain, Ravid Shwartz-Ziv, Neel Jain, Khalid Saifullah, Siddartha Naidu, et~al.
\newblock Livebench: A challenging, contamination-free llm benchmark.
\newblock \emph{arXiv preprint arXiv:2406.19314}, 4, 2024.

\bibitem[Xu et~al.(2025)Xu, Hao, Zong, Wang, Zhang, Wang, Lan, Gong, Ouyang, Meng, et~al.]{xu2025towards}
Fengli Xu, Qianyue Hao, Zefang Zong, Jingwei Wang, Yunke Zhang, Jingyi Wang, Xiaochong Lan, Jiahui Gong, Tianjian Ouyang, Fanjin Meng, et~al.
\newblock Towards large reasoning models: A survey of reinforced reasoning with large language models.
\newblock \emph{arXiv preprint arXiv:2501.09686}, 2025.

\bibitem[Yadav et~al.(2024)Yadav, Tam, Choshen, Raffel, and Bansal]{yadav2024ties}
Prateek Yadav, Derek Tam, Leshem Choshen, Colin Raffel, and Mohit Bansal.
\newblock Ties-merging: Resolving interference when merging models.
\newblock \emph{Advances in Neural Information Processing Systems}, 36, 2024.

\bibitem[Yu et~al.(2025)Yu, Zhang, Zhu, Yuan, Zuo, Yue, Dai, Fan, Liu, Liu, et~al.]{yu2025dapo}
Qiying Yu, Zheng Zhang, Ruofei Zhu, Yufeng Yuan, Xiaochen Zuo, Yu~Yue, Weinan Dai, Tiantian Fan, Gaohong Liu, Lingjun Liu, et~al.
\newblock Dapo: An open-source llm reinforcement learning system at scale.
\newblock \emph{arXiv preprint arXiv:2503.14476}, 2025.

\bibitem[Yue et~al.(2025)Yue, Yuan, Yu, Zuo, Zhu, Xu, Chen, Wang, Fan, Du, et~al.]{yue2025vapo}
Yu~Yue, Yufeng Yuan, Qiying Yu, Xiaochen Zuo, Ruofei Zhu, Wenyuan Xu, Jiaze Chen, Chengyi Wang, TianTian Fan, Zhengyin Du, et~al.
\newblock Vapo: Efficient and reliable reinforcement learning for advanced reasoning tasks.
\newblock \emph{arXiv preprint arXiv:2504.05118}, 2025.

\bibitem[Zheng et~al.(2024{\natexlab{a}})Zheng, Sun, Wu, Xi, and Zhou]{zheng2024balancing}
Chen Zheng, Ke~Sun, Hang Wu, Chenguang Xi, and Xun Zhou.
\newblock Balancing enhancement, harmlessness, and general capabilities: enhancing conversational llms with direct rlhf.
\newblock \emph{arXiv preprint arXiv:2403.02513}, 2024{\natexlab{a}}.

\bibitem[Zheng et~al.(2024{\natexlab{b}})Zheng, Sun, and Zhou]{zheng2024mistral}
Chen Zheng, Ke~Sun, and Xun Zhou.
\newblock Mistral-c2f: Coarse to fine actor for analytical and reasoning enhancement in rlhf and effective-merged llms.
\newblock \emph{arXiv preprint arXiv:2406.08657}, 2024{\natexlab{b}}.

\end{thebibliography}

\clearpage

\beginappendix


\section{Detailed Mean Response Length across Different BAI Ratio}
\label{sec:mean_response_len}

\section{Reward Score Curve across Different BAI Ratio}

\section{Case Study}

\begin{table}[h]
\centering
\footnotesize
\begin{tabular}{|p{2cm}|p{12cm}|}
\hline
Prompt 

Label: A
& According to Stevenson, moral disagreements involve:

A) disagreement in interest. 

B) disagreement in belief. 

C) disagreement in theory. 

D) disagreement in practice. \\
\hline
Paradigm 1 & 1. Analyze each option:

Option A: Disagreement in interest typically refers to differences in what people want or desire. In the context of moral disagreements, moral issues are not primarily about different interests. For example, if two people are arguing about whether it is wrong to lie to save a life, it is not mainly a matter of different interests they have. So, option A is incorrect.

Option B: Stevenson argues that moral disagreements involve disagreement in belief. When people have moral disagreements, they often have different beliefs about what is right and wrong. For instance, one person may believe that euthanasia is always wrong based on religious beliefs, while another person may believe that under certain circumstances, euthanasia is morally acceptable. This difference is a disagreement in belief about the moral status of euthanasia. So, option B is correct.

Option C: Disagreement in theory is too broad and not the core aspect of what Stevenson emphasizes in moral disagreements. The focus is on the underlying beliefs about moral values rather than differences in theoretical frameworks in general. So, option C is incorrect.

Option D: Disagreement in practice may seem related, but the key aspect of moral disagreements according to Stevenson is the underlying disagreement in beliefs about what is morally right or wrong, not just differences in how people act in practice. So, option D is incorrect.

Answer: B \\
\hline
Paradigm 2 & To determine which option Stevenson identifies as involved in moral disagreements, let's recall Stevenson's views on ethical disagreement. Stevenson argued that moral disagreements primarily stem from differences in attitudes rather than just differences in beliefs.

He distinguished between two types of disagreement: disagreement in belief (where people disagree about the facts of a situation) and disagreement in attitude (which involves differing emotional or evaluative stances). In moral disputes, Stevenson contended that the key source of conflict is not merely about factual beliefs but about conflicting desires, preferences, or values.

Option A, "disagreement in interest," might seem related, but Stevenson's focus was more on the clash of attitudes or interests rather than a specific type of interest. Option C, "disagreement in theory," is too broad and not specific to moral disagreements. Option D, "disagreement in practice," refers to actions but does not capture the underlying disagreement in attitudes that Stevenson emphasized.

Option B, "disagreement in belief," aligns with Stevenson's view that moral disagreements often involve differing factual beliefs about the consequences or nature of an action. However, he also stressed that the deeper disagreement is in attitudes. But given the options provided, the most accurate according to Stevenson's analysis would be that moral disagreements involve disagreement in belief, even though there's a broader element of disagreement in attitude.

Answer: B \\
\hline
Paradigm 3 & Robert L. Stevenson was a philosopher known for his work on emotivism, a meta-ethical theory that posits moral statements primarily express emotions or attitudes rather than objective facts. In his view, moral disagreements do not stem from differences in beliefs about factual matters.

Emotivists like Stevenson argue that when people have moral disagreements, the core issue isn't that one person has a belief that the other lacks. Instead, the disagreement arises from differing attitudes or emotional responses towards a particular action or situation. For instance, if one person thinks an action is right and another thinks it's wrong, this isn't because they have different beliefs about the action's consequences or properties; rather, it's because they have contrasting emotional reactions towards it.

Thus, moral disagreements, according to Stevenson, involve disagreement in interest—different emotional or attitudinal stances towards the same object.

Options B, C, and D don't capture the essence of Stevenson's position. Disagreement in belief (B) implies a difference in factual understanding, which emotivism rejects as the primary cause of moral disputes. Disagreement in theory (C) refers to overarching philosophical frameworks, which isn't the main focus here. Disagreement in practice (D) would suggest differing application of principles, but Stevenson's view is that the disagreement is at the level of emotional response, not practical application.

Answer: A \\
\hline
\end{tabular}
\caption{Case Study: Stevenson's Theory of Moral Disagreements}
\end{table}

\begin{table}[h]
\centering
\footnotesize
\begin{tabular}{|p{2cm}|p{12cm}|}
\hline
Prompt & I would like to build a magic the gathering deck. this deck contains 99 cards. How many lands should i play to draw on average 3 lands in my starting 7 cards hand. explain the calculus.\\
\hline
Paradigm 1 & 1. Use the hypergeometric distribution formula: In a Magic: The Gathering deck, assume the total number of cards in the deck is $N = 99$ (the size of the deck). Let the number of lands in the deck be $K$. The number of cards drawn from the deck is $n = 7$ (the starting hand size). We want the expected number of lands in the hand to be $E(X)=3$.

The probability mass function of the hypergeometric distribution for the number of successes (lands in this case) in a sample of size $n$ drawn from a population of size $N$ containing $K$ successes is given by $P(X = k)=\frac{\binom{K}{k}\binom{N - K}{n - k}}{\binom{N}{n}}$, where $\binom{a}{b}=\frac{a!}{b!(a - b)!}$. The expected value $E(X)$ of a hypergeometric random variable $X$ is given by the formula $E(X)=n\times\frac{K}{N}$.

2. Solve for the number of lands $K$: We know that $n = 7$, $N = 99$, and $E(X)=3$. Substituting these values into the formula $E(X)=n\times\frac{K}{N}$, we get: $3=7\times\frac{K}{99}$. First, we can rewrite the equation as $3=\frac{7K}{99}$. To solve for $K$, we multiply both sides of the equation by $99$: $3\times99 = 7K$. So, $297 = 7K$. Then, we divide both sides by $7$ to find $K=\frac{297}{7}\approx42.43$.

Since we can't have a fraction of a land in the deck, we would typically play 42 lands in the 99-card deck to get, on average, 3 lands in the starting 7-card hand. \\
\hline
Paradigm 2 & The Gathering deck to have an average of 3 lands in your starting 7-card hand, we need to use probability and combinatorics.

\textbf{Key Concepts:} 1. Hypergeometric Distribution: This models the probability of drawing a certain number of "successes" (lands in this case) from a finite population without replacement. 2. Average (Expected Value): The expected number of lands in a 7-card hand can be calculated using the hypergeometric distribution formula.

\textbf{Formula for Expected Value (Mean) of a Hypergeometric Distribution:} The expected number of lands $E(X)$ in a 7-card hand, given a deck of $N$ cards with $K$ lands, is: $E(X) = \frac{7 \cdot K}{N}$. We want $E(X) = 3$. Given $N = 99$ cards, we can solve for $K$: $3 = \frac{7 \cdot K}{99}$. Solving for $K$: $K = \frac{3 \cdot 99}{7} \approx \frac{297}{7} \approx 42.43$. Since we can't include a fraction of a land, we round up to 43 lands.

\textbf{Check with a Smaller Example:} Let's verify with a smaller deck (e.g., 10 cards with 4 lands): Expected lands in a 3-card hand: $\frac{3 \cdot 4}{10} = 1.2$. Similarly, in a 7-card hand from 99 cards with 43 lands: $E(X) = \frac{7 \cdot 43}{99} \approx \frac{301}{99} \approx 3.04$, which is approximately 3 lands.

\textbf{Conclusion:} To have an average of 3 lands in your starting 7-card hand, include 43 lands in your 99-card deck. This ensures that the expected value aligns with your goal, though individual hands will vary due to randomness. \\
\hline
Paradigm 3 & To determine how many lands you should include in a 99-card Magic: The Gathering deck to have an average of 3 lands in your starting 7-card hand, we can use the concept of probability and expected value.

\textbf{The Basic Idea:} The probability of drawing exactly $k$ lands in a 7-card hand from a deck containing $n$ lands out of 99 total cards can be calculated using the hypergeometric distribution. The expected value (average) of the number of lands in the starting hand is given by: $E(X) = \frac{7 \times n}{99}$. We want $E(X) = 3$, so we can set up the equation: $\frac{7 \times n}{99} = 3$.

\textbf{Solving for $n$:} Let's solve for $n$: $7n = 3 \times 99$, $7n = 297$, $n = \frac{297}{7} \approx 42.4286$. Since the number of lands must be an integer, we round up to the nearest whole number.

\textbf{Result:} To have an average of 3 lands in your starting 7-card hand in a 99-card deck, you should include 43 lands. Verification: $\frac{7 \times 43}{99} = \frac{301}{99} \approx 3.0404$. This is close to 3, meeting the desired average.

\textbf{Why the Hypergeometric Distribution?} The hypergeometric distribution is appropriate here because: 1. We are drawing a fixed sample size (7 cards) without replacement from a finite population (99 cards). 2. There are two distinct groups: lands and non-lands. The formula for the probability of drawing $k$ lands is: $P(X = k) = \frac{\binom{n}{k} \binom{99 - n}{7 - k}}{\binom{99}{7}}$ where $\binom{a}{b}$ is the binomial coefficient (number of ways to choose $b$ items from $a$). The expected value $E(X)$ simplifies to $\frac{7n}{99}$, as mentioned earlier.

\textbf{Conclusion:} In a 99-card Magic deck, including 43 lands will give you an average of approximately 3 lands in your starting 7-card hand. Adjustments can be made based on your playstyle (e.g., card draw or mana ramp), but 43 is a solid starting point. \\
\hline
\end{tabular}
\caption{Case Study: Magic the Gathering Deck Land Calculation}
\end{table}

\end{document}